\documentclass[10pt,twocolumn,letterpaper]{article}

\usepackage{cvpr}              
\usepackage{algorithm}       
\usepackage{algorithmicx}    
\usepackage{algpseudocode}   

\usepackage{multirow}
\usepackage{graphicx}
\usepackage{subcaption}
\usepackage[accsupp]{axessibility}
\definecolor{cvprblue}{rgb}{0.21,0.49,0.74}
\usepackage[pagebackref,breaklinks,colorlinks,allcolors=cvprblue]{hyperref}

\def\paperID{} 
\def\confName{CVPR}
\def\confYear{2026}

\title{Residual Decoding: Mitigating Hallucinations in Large Vision-Language Models via History-Aware Residual Guidance}

\author{
  \textbf{Xinrong Chen}\textsuperscript{1}\thanks{\ Equal contribution.} \quad
  \textbf{Xu Chu}\textsuperscript{1}\footnotemark[1] \quad
  \textbf{Yingmin Qiu}\textsuperscript{2}\footnotemark[1] \quad
  \textbf{Hengyuan Zhang}\textsuperscript{3}\thanks{\ Corresponding author.} \quad
  \textbf{Jing Xiong}\textsuperscript{3} \quad 
  \textbf{Shiyu Tang}\textsuperscript{4} \\
  \textbf{Shuai Liu}\textsuperscript{4} \quad
  \textbf{Shaokang Yang}\textsuperscript{4} \quad
  \textbf{Cheng Yang}\textsuperscript{4} \quad
  \textbf{Hayden Kwok-Hay So}\textsuperscript{3} \quad
   \textbf{Ngai Wong}\textsuperscript{3}\footnotemark[2]  \\ 
   \textsuperscript{1}Peking University 
  \textsuperscript{2}Beijing University of Posts and Telecommunications \\
   \textsuperscript{3}University of Hong Kong
  \textsuperscript{4}ByteDance Inc. 
   \\
\texttt{chenxinrong23@stu.pku.edu.cn\ \  chuxu@stu.pku.edu.cn}  
}

\begin{document}
\maketitle
\begin{abstract}
Large Vision-Language Models (LVLMs) can reason from image-text inputs and perform well in various multimodal tasks. Despite this success, they are affected by language priors and often produce hallucinations. Hallucinations denote generated content that is grammatically and syntactically coherent, yet bears no match or direct relevance to visual input. To address this problem, we propose Residual Decoding (\textit{ResDec}). It is a novel training-free method that uses historical information to aid decoding. The method relies on the internal implicit reasoning mechanism and token logits evolution mechanism of LVLMs to correct biases. Extensive experiments demonstrate that \textit{ResDec} effectively suppresses hallucinations induced by language priors, significantly improves visual grounding, and reduces object hallucinations. In addition to mitigating hallucinations, \textit{ResDec} also performs exceptionally well on comprehensive LVLM benchmarks, highlighting its broad applicability.
\end{abstract}    
\section{Introduction}
\label{sec:intro}
Large Vision-Language Models (LVLMs)~\cite{gong2023multimodalgptvisionlanguagemodel,zhu2025internvl3exploringadvancedtraining,dai2023instructblipgeneralpurposevisionlanguagemodels,chang2025treereview,Liu_2024_CVPR,zhang2025guilomo,bai2025qwen25vltechnicalreport,xiong2026mmformalizer,alayrac2022flamingovisuallanguagemodel,liang2026divideconquer,shang2024incremental,shang2024understanding,vteam2025glm45vglm41vthinkingversatilemultimodal,chu2025domaino1s,yuan2026more} have become a core component of modern artificial intelligence, playing a crucial role in various vision-language tasks by seamlessly combining visual perception and language understanding. Despite significant progress in recent years, LVLMs are plagued by hallucinations~\cite{gunjal2024detectingpreventinghallucinationslarge,li2023evaluatingobjecthallucinationlarge,huang-etal-2024-visual,tan2024order,zheng-etal-2025-reefknot,zhang2024balancing,liu2024mitigatinghallucinationlargemultimodal,chen2025surveymultimodalhallucinationevaluation}, which impede their reliability and applicability in tasks.

Hallucinations in LVLMs can be defined as generating content that is irrelevant to or contradicts the facts in the image~\cite{favero2024multi}. Many studies show that LVLMs hallucinate for three main reasons: statistical pre-training bias~\cite{agarwal2020causalvqarevealingreducing,agrawal-etal-2016-analyzing,li2025analyzingmitigatingobjecthallucination}, weak vision–language alignment~\cite{zhou2024calibratedselfrewardingvisionlanguage,zhao2025mitigatingobjecthallucinationlarge,zhu2025mitigatingobjecthallucinationslarge}, and optimization bias~\cite{fu2024mitigatinghallucinationmultimodallarge,compagnoni2025mitigatinghallucinationsmultimodalllms,liu2025modalitybalancingpreferenceoptimizationlarge}. In addition, accumulated language priors~\cite{favero2024multi, zhou2023analyzing, li2025hidden,chu2025qwen} during generation further exacerbate these hallucinations. \emph{That is, when the model generates responses, textual content gradually fades the visual context, leading to grammatically coherent but visually ungrounded content.} This phenomenon can be termed Language-Prior Hallucination.

\begin{figure}[t]
  \centering
   \includegraphics[width=1.0\linewidth]{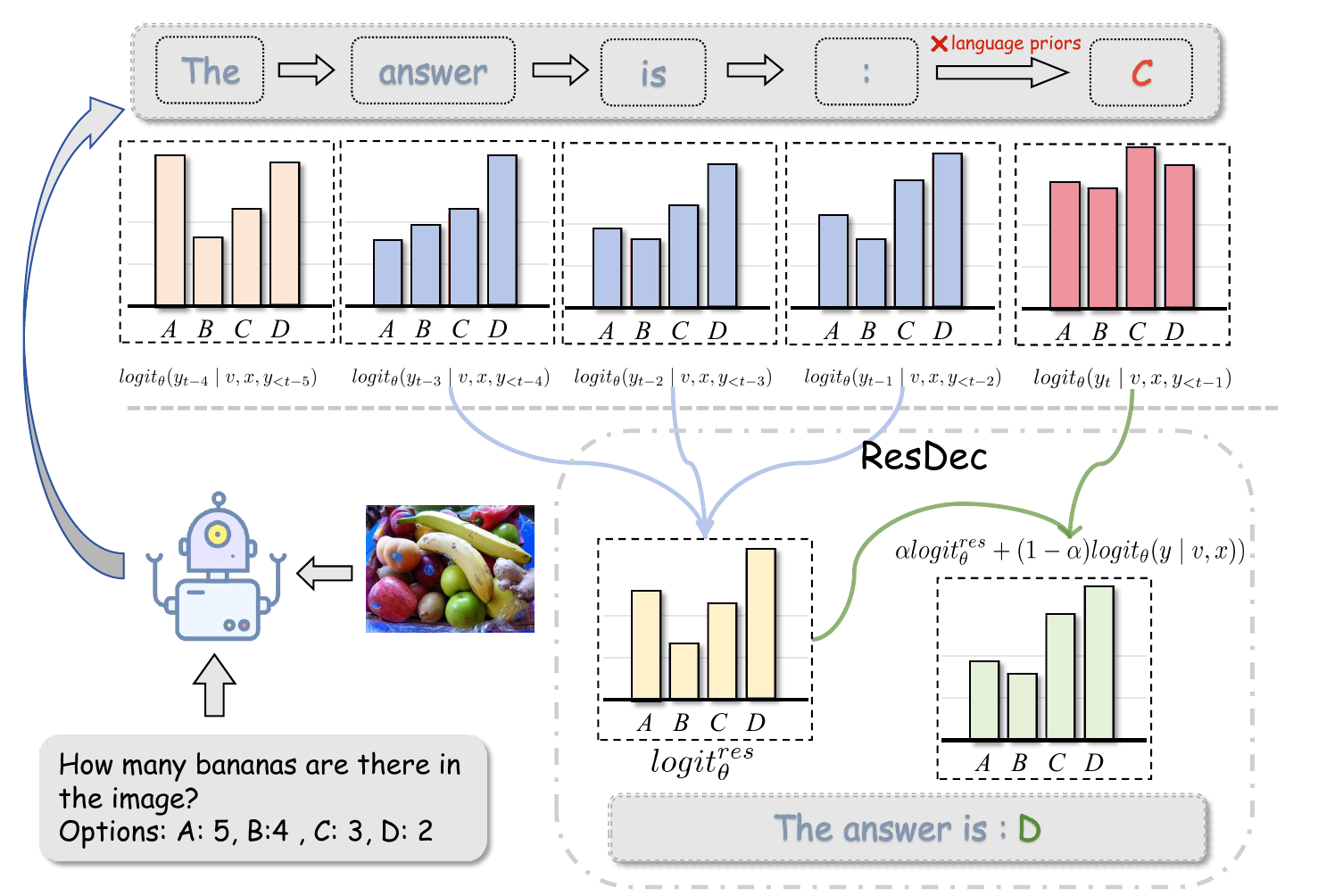}

   \caption{An overview of Residual Decoding. The figure illustrates the Language-Prior Hallucination by focusing on candidate tokens (i.e. A, B, C, and D), and indicates that Residual Decoding with residual guidance effectively mitigates such hallucinations.}
   \label{fig:intro}
   \vspace{-0.5cm}
\end{figure}

Researchers have explored various approaches to mitigate Language-Prior Hallucination in LVLMs. Training-based methods include data debiasing~\cite{peng2025mitigatingobjecthallucinationssentencelevel,zhang2025debiasingmultimodallargelanguage,sun2023aligninglargemultimodalmodels}, modality alignment enhancement~\cite{zhang2025robustmultimodallargelanguage,jiang2024hallucinationaugmentedcontrastivelearning,jiang2024hallucinationaugmentedcontrastivelearning}, and preference alignment~\cite{peng2025mitigatingobjecthallucinationssentencelevel,wu2025mitigatinghallucinationslargevisionlanguage,yang2025mitigatinghallucinationslargevisionlanguage}. However, these methods require additional training and costly annotations, which limit their scalability in practice. Another line of research focuses on training-free methods, aiming to alleviate hallucinations in a cost-effective manner. For example, recent studies propose contrastive decoding strategies~\cite{leng2023mitigatingobjecthallucinationslarge,wang2024mitigatinghallucinationslargevisionlanguage,an2025mitigatingobjecthallucinationslarge} that contrast model outputs under perturbed visual or textual conditions to suppress language priors. Other studies intervene within the model architecture, such as in attention mechanisms~\cite{wan2025onlyonelayerinterventionsufficiently,huang2024operaalleviatinghallucinationmultimodal,zhang2026locatesteersurvey,liu2024payingattentionimagetrainingfree}, feed-forward networks (FFNs)~\cite{zou2025looktwiceanswermemoryspace}, or at the layer level~\cite{guo2025lisalayerwiseintegrationsuppression,chuang2023dola,li2025hiddenlifetokensreducing}, to enhance visual perception and reduce hallucinations in LVLMs. Despite their effectiveness, these training-free methods still have notable limitations:
\begin{enumerate*}[label=(\roman*)]
\item they require $2\times$ or even more inference time and incur higher GPU memory overhead.
\item they modify internal components of the model, making them less efficient, less concise, and potentially less generalizable.
\end{enumerate*}
To develop a simple, efficient, and highly generalizable method for mitigating Language-Prior Hallucination, a feasible solution is to design a decoding strategy that is less susceptible to language priors. To achieve this, we revisit the decoding process of LVLMs and find that \textbf{signals of the correct answer are already embedded in the logit distributions of preceding tokens before the model explicitly decodes the answer}. As shown in Fig.\ref{fig:intro}, when the model sequentially generates guiding tokens such as ``The'', ``answer'', ``is'', the logits of the correct answer token ``D'' are already at relatively high values in the logits of these historical tokens. However, when generating the token ``:'', the model assigns an abnormally high logits value to the hallucinated token ``C'', causing its probability score to surpass that of the genuine token ``D''. Ultimately, when the model explicitly decodes the answer, it outputs the incorrect ``C''. This observation reveals the essential mechanism of hallucinations in LVLMs: \textbf{hallucinated tokens erroneously emerge in the logit distribution at certain moments, and their probabilities gradually approach or even exceed those of genuine tokens}, thereby causing the model to produce hallucinations during final decoding.

To address this issue, we propose \textbf{Residual Decoding (\textit{ResDec})}, a training-free and plug-and-play decoding strategy. \textit{ResDec} first utilizes Jensen–Shannon Divergence (JSD) to analyze the logits of historical tokens within a time window and discovers a U-shaped pattern (see \S~\ref{sec:imp_infer}). On the left side of the JSD valley, the distribution transitions from ``chaotic'' to ``convergent''. Near the valley, the distribution becomes smooth and the JSD between adjacent time steps is small, 
the distribution becomes smooth and the JSD between adjacent time steps is small, indicating a stable logits distribution. This stability suggests that semantics have largely converged and predictions are less influenced by language priors. On the right side of the valley, the distribution diverges again as expressions become divergent. Accordingly, \textit{ResDec} selects logits from the post-convergence region (i.e., near and to the right of the JSD valley), since time steps in this region have stable semantics and exert a substantial effect on the output distribution. These selected logits are then aggregated with confidence weights, yielding a residual guidance that is fused into the decoding logits to reduce Language-Prior Hallucination. \textit{ResDec} requires only a single forward pass, without any contrastive model or external encoder, and incurs nearly the same inference cost as standard decoding. Extensive experiments on multiple LVLM hallucination benchmarks demonstrate that \textit{ResDec} achieves state-of-the-art performance in reducing model hallucinations. Our contributions can be summarized as follows:

\begin{itemize}
\item We discover that the essence of hallucinations in LVLMs is that hallucinated tokens emerge in the logits during the decoding stage and surpass genuine tokens (see Fig.~\ref{fig:intro}).
\item We propose \textbf{Residual Decoding (\textit{ResDec})}, an efficient, training-free decoding strategy that incurs negligible inference cost for mitigating hallucinations. \textit{ResDec} uses JSD to analyze the logit distributions of historical tokens, aggregates logits from semantically clear tokens to form a residual stream, and merges it into the logits of the decoded token, suppressing Language-Prior Hallucination.
\item Extensive experiments reveal that \textit{ResDec} reduces hallucinations in LVLMs and yields improvements on comprehensive datasets, highlighting its broad generalization capability. For instance, \textit{ResDec} achieves an average improvement of $7.84\%$ in accuracy and $8.01\%$ in F1 over the ``Regular'' decoding strategy across three LVLMs.
\end{itemize}

\section{Related Work}
\subsection{Causes of Hallucinations in LVLMs}
Hallucination~\cite{zhang2025robustmultimodallargelanguage,peng2025mitigatingobjecthallucinationssentencelevel,liang2025sws,gunjal2024detectingpreventinghallucinationslarge,li2023evaluatingobjecthallucinationlarge,huang-etal-2024-visual,zheng-etal-2025-reefknot,zhang2025find,liu2024mitigatinghallucinationlargemultimodal,chen2025surveymultimodalhallucinationevaluation} has long been a major challenge in Large Vision-Language Models (LVLMs). 
Recent studies attribute this phenomenon to various factors, such as statistical pre-training bias~\cite{ramos2025globallocalsocialbias,huang2025shieldsuppressinghallucinationslvlm,li2025analyzingmitigatingobjecthallucination}, weak vision–language alignment~\cite{zhang2025robustmultimodallargelanguage,jiang2024hallucinationaugmentedcontrastivelearning,zhang-etal-2025-shifcon,jiang2024hallucinationaugmentedcontrastivelearning}, and optimization objectives~\cite{yang2025mitigatinghallucinationslargevisionlanguage} that favor linguistic coherence over factual grounding. Notably, \textit{language priors}~\cite{favero2024multi, zhou2023analyzing, li2025hidden,chu2025qwen} that further exacerbate hallucinations induced by the aforementioned factors have emerged as the most precisely defined and extensively studied cause.
Several studies demonstrate that generative VLMs can, in some cases, completely ignore visual inputs while still outperforming previous approaches, highlighting the dominant influence of language priors~\citep{lin2024revisitingrolelanguagepriors,zhao2024hallucinationsenhancinglvlmshallucinationaware,jiang2024hallucinationaugmentedcontrastivelearning}. Likewise, \citet{lee-etal-2025-vlind} introduces a benchmark specifically designed to quantify the extent to which models rely on language priors independent of visual evidence. These studies show that language priors can bias LVLMs towards generating linguistically plausible but visually ungrounded outputs, leading to hallucinations.

\subsection{Mitigating Language-Prior Hallucination}
To address hallucinations driven by language priors in LVLMs, 
\citet{wang2024mitigatinghallucinationslargevisionlanguage} contrasted output distributions under varied prompts to reduce hallucinated tokens. Moreover, \citet{min2025mitigatinghallucinationslargevisionlanguage} encouraged the model to prioritize image input by progressively condensing textual context, thereby reducing reliance on linguistic patterns.
\citet{wan2025onlyonelayerinterventionsufficiently} proposes modifying attention weights to mitigate the impact of textual biases and enhance visual grounding.
\citet{yang2025mitigatinghallucinationslargevisionlanguage} employed on-policy human preference data with Direct Preference Optimization (DPO) to refine the generation strategy, reducing the influence of language priors and improving visual grounding.
However, these methods face inherent limitations, including additional computational costs, dependence on extra fine-tuning data, and the need for architectural interventions, which compromises their simplicity, efficiency, and generalizability.
Against this backdrop, we propose \textit{ResDec}, a residual decoding strategy tailored to mitigate Language-Prior Hallucination while addressing the aforementioned limitations.

\section{Methodology}
\label{sec:method}

\subsection{Problem Formulation}
\paragraph{LVLM Decoding.}
Consider an LVLM parameterized by $\theta$, with a general architecture consisting of a vision encoder, a vision–language connector, and a Large Language Model (LLM). 
Given an image $v$ and query ${x}$, the LVLM first encodes $v$ into visual tokens using the vision encoder, and then concatenates them with the text tokens from ${x}$. 
The combined sequence is then fed into the LLM to generate the output sequence $y$ autoregressively:
\begin{equation}
y_t \sim p_\theta(y_t \mid v, {x}, {y}_{<t})
= \mathrm{softmax}(\text{logit}_\theta(y_t \mid v, {x}, {y}_{<t})),
\label{eq:lvlm_decoding}
\end{equation}
\noindent where $y_t$ denotes the $t$-th generated token, ${y}_{<t}$ is the sequence of previously generated tokens, $p_\theta$ represents the conditional probability distribution parameterized by the LVLM, and $\text{logit}_\theta$ denotes the logit distribution.

\paragraph{Language Prior.}
\label{sec:lp}
Taking the mathematical interpretation from \citet{lin2024revisitingrolelanguagepriors}, we view the \emph{language prior} as the text-only, vision-agnostic conditional distribution $P(y \mid x)$, i.e., the distribution over outputs $y$ given the textual context $x$ when the visual input $v$ is removed. 
By considering two equivalent factorizations of $P(y, v \mid x)$ and applying Bayes rule, one can show that the \emph{language prior} can be ``divided out'' from the visually conditioned prediction:
\begin{equation}
\frac{P(y \mid v, x)}{P(y \mid x)}
= \frac{P(v \mid y, x)}{P(v \mid x)}.
\label{eq:pmi_ratio_main}
\end{equation}
\noindent This ratio is analogous to pointwise mutual information (PMI~\cite{role2011handling}) and captures how the image $v$ supports an output $y$ beyond the language prior. 
To further make the language prior explicit, it can be expressed as a marginalization over the visual variable and approximated via Monte Carlo~\cite{SHAPIRO2003353}:
\vspace{-0.2cm}
{
\small
\begin{equation}
P(y \mid x)
=
\mathbb{E}_{v \sim P(v \mid x)}\!\big[P(y \mid v, x)\big]
\;\approx\;
\frac{1}{n}\sum_{j=1}^{n} P\!\big(y \mid v_j, x\big),
\label{eq:language_prior_mc}
\end{equation}
}

\noindent where $\{v_j\}_{j=1}^{n}$ are samples from the image distribution conditioned on $x$. We adopt this probabilistic view and provide the full derivation in Appendix~\ref{appendix:language_priors}.

\subsection{Implicit Inference of LVLMs}
\label{sec:imp_infer}

Recent research reveals that LLMs possess implicit multi-token prediction capabilities~\cite{pal-etal-2023-future, samragh2025llmknowsfutureuncovering}, which means the model already ``knows'' part of its future outputs before actual decoding. 
Specifically, as illustrated in Fig.~\ref{fig:intro}, the high-ranking token candidates in the logits at time step $t$ are already implicitly contained in the logits generated during the production of the previous token sequence from $y_{t-4}$ to $y_{t-1}$. This phenomenon provides a potential avenue for mitigating Language-Prior Hallucination in LVLMs, namely by analyzing the high-ranking tokens in the logits of $y_{<t}$ to help decode correct content at time step $t$. To further understand this phenomenon and achieve the goal of mitigating hallucinations, we propose two questions:

\begin{itemize}
    \item \textbf{RQ1}: As the input context evolves, can the LVLM progressively and implicitly represent the answer?
    \item \textbf{RQ2}: How do the high-ranking candidate tokens at timestep $t$ change in the logits at timestep $<t$?
\end{itemize}

\begin{figure}[t]
  \centering
   \includegraphics[width=1.0\linewidth]{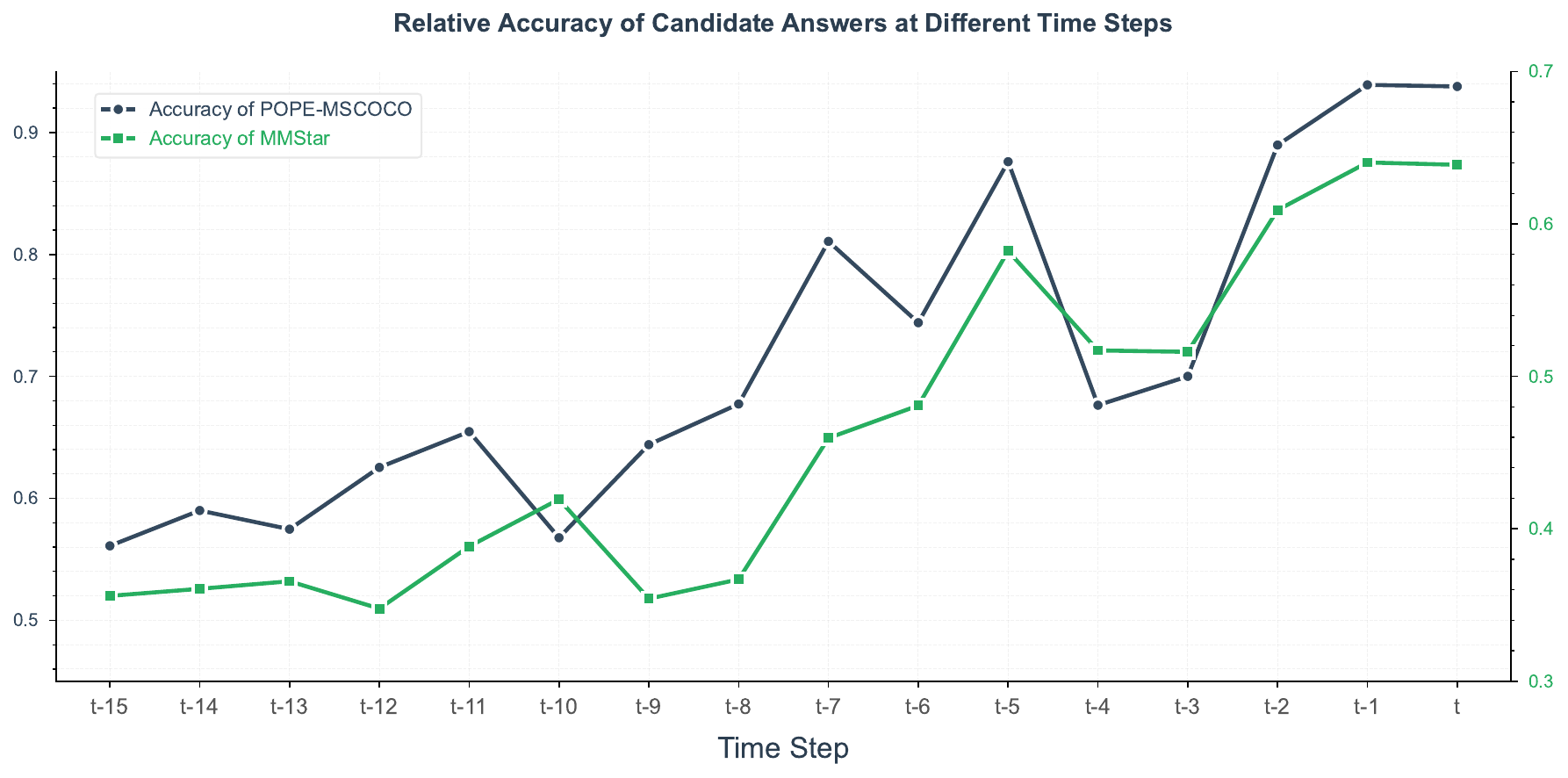}

   \caption{Accuracy of candidate answers at different time steps based on Qwen2.5-VL (7B) in MME and POPE. In both datasets, the candidate answers are binary \textit{Yes/No} decisions.}
   \label{fig:rq1}
   \vspace{-0.2cm}
\end{figure}

\noindent \textbf{Impact of Context Evolution on Implicit Inference (RQ1).}
As shown in Fig.~\ref{fig:rq1}, we observe that as the input context evolves, especially as the question is gradually described completely, the LVLM's forward-looking perception of the correct reasoning trajectory is progressively activated, manifested as an increase in prediction accuracy and reaching a peak near the token to be decoded (i.e., near the right end of the coordinate axis). This indicates that when the question approaches complete description, the model's logits already implicitly represent the answer. However, we observe that implicit representation is not always stable. We attribute this to the fact that during the step-by-step token generation process, the LVLM gradually accumulates two different types of biases. Specifically, 1) \textit{Natural language grammar preferences} require the decoded content to satisfy grammar, word order, and punctuation from the pre-training phase. 2) \textit{Task-specific template preferences} require the decoded content to satisfy dialogue templates from the instruction fine-tuning phase (e.g., response templates such as ``The answer is :''). This causes the logits of implicit inference to contain not only the answer but also to be affected by other noise. To leverage implicit inference for hallucination mitigation, it is necessary to identify stable implicit representation phases.

\begin{figure}[t]
  \centering
   \includegraphics[width=1.0\linewidth]{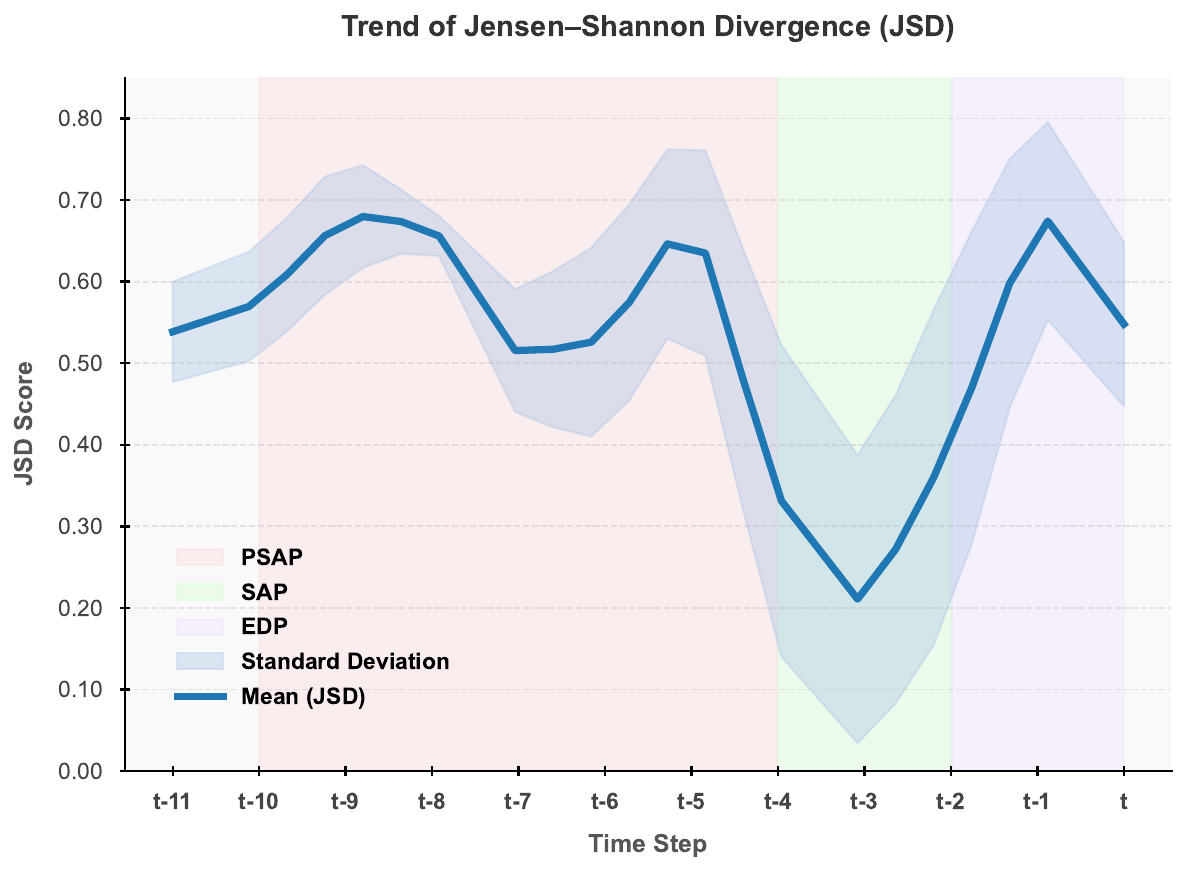}
    
   \caption{The trend of Jensen–Shannon Divergence (JSD) on $200$ randomly sampled POPE-MSCOCO examples using Qwen2.5-VL-Instruct (7B). JSD exhibits a distinct U-shape over time, which we divide into three phases: \textbf{PSAP}, \textbf{SAP}, and \textbf{EDP}. 
 }
   \label{fig:rq2}
   \vspace{-0.4cm}
\end{figure}

\noindent \textbf{Temporal Evolution of Candidate Tokens (RQ2).}
In this section, our core focus is to explore the fine-grained temporal evolution patterns of the probability distributions of candidate tokens (including genuine tokens and hallucinated tokens). We first define the targeted candidate token set $\Omega$, which consists of the top-$k$ tokens with the highest probabilities at a given time step $t$. Then, we use a sliding historical window $\mathcal{W}$ to limit temporal scope for focused analysis. Within this window, we calculate the Jensen–Shannon Divergence (JSD~\cite{Bri_t_2009}) between the probability distributions of $\Omega$ at each adjacent pair of time steps: 

\vspace{-0.3cm}
{
\small
\begin{equation}
\text{JSD}(P_{t_1} \parallel P_{t_2}) = \frac{1}{2} D_{\text{KL}}(P_{t_1} \parallel M) + \frac{1}{2} D_{\text{KL}}(P_{t_2} \parallel M),
\label{eq:jsd}
\end{equation}
}

\noindent where $M = \frac{1}{2}(P_{t_1}+P_{t_2})$ and $D_{\text{KL}}$ denotes the Kullback–Leibler (KL) divergence~\cite{kullback1951information}. Specifically, we compute the JSD curves by averaging over $200$ randomly sampled examples from POPE-MSCOCO. As shown in Fig.~\ref{fig:rq2}, we observe that JSD values exhibit a U-shaped trend over time. Taking the minimum point of the U-shape as the boundary, the temporal evolution of the candidate token distribution can be decomposed into three phases:

\begin{itemize}[leftmargin=2em]
    \item[\textbullet] \textbf{(Phase-1) Pre-Semantic Clarity Phase (PSAP)}: From the onset to the bottom of the U-shape, JSD exhibits a fluctuating downward trend. We interpret this as indicating that the candidate token distribution is transitioning from an initial state of ``disorder'' to a state of ``convergence''. The minor oscillations in JSD within this phase reflect temporary uncertainty in semantic anchoring, as the model iteratively narrows down its candidate token choices while still navigating potential ambiguities before settling on the core semantics.

    \item[\textbullet] \textbf{(Phase-2) Semantic Anchoring Phase (SAP)}: At the bottom of the U-shape, JSD approaches $0$. This near-zero JSD indicates that the candidate token distribution stabilizes to its highest degree, with minimal divergence between consecutive time steps, reflecting the LVLM's firm anchoring to core semantics.
    
    \item[\textbullet] \textbf{Expressive Divergence Phase (EDP)}: From the bottom of the U-shape to time step $t$, JSD shows an upward trend, as the LVLM strives to align with the contextual template and pursues diverse expressive forms for candidate tokens. Notably, this phase is susceptible to language priors, as the pursuit of varied expressions may increase reliance on linguistic patterns, causing deviations from visual or factual semantics.
\end{itemize}

\textbf{PSAP} clearly belongs to an unstable representation phase. Compared to \textbf{SAP}, \textbf{EDP} appears less stable, it still potentially contains effective decoding guidance due to its incorporation of diverse expressions. Based on the above observations and analyses, we propose \textit{ResDec}, which guides token selection through historical information aggregation, thereby stabilizing the decoding process.

\subsection{Residual Decoding}
In this section, we present the two key components of \textit{ResDec} strategy: 1) Confidence-weighted aggregation of historical information, discussed in \S~\ref{sec:hig}; and 2) Integration of historical and current information, elaborated in \S~\ref{sec:ahp}.
\subsubsection{Historical Information Aggregation}
\label{sec:hig}
Historical Information Aggregation is a core component of \textit{ResDec}, with the primary goal of dynamically capturing and reusing historical contextual information from the decoding process, which aims to provide sustained and valid references for subsequent token selection.

\vspace{0.1cm}
\noindent\textbf{Historical Aggregation Window.}
As demonstrated in \S~\ref{sec:imp_infer}, meaningful historical windows often reside in \textbf{SAP} and \textbf{EDP} phases. To obtain this window, we typically need to retrieve the corresponding steps from the Semantic Anchoring Phase. This requires empirically setting a predefined historical window $W$, which will be discussed in Appendix~\ref{appendix:window}, and then selecting the high-rank tokens at the current time step 
$t$ as Candidate Token Pool $\Omega_t$: 

\vspace{-0.4cm}
\begin{equation}
\Omega_t = \left\{ y_t \mid y_t \in \text{top-}k \left( \text{logit}_\theta(y_t \mid v, x, y_{<t}), K\right) \right\},
\label{eq:candidate_pool}
\end{equation}
\noindent Given that \textbf{SAP} and \textbf{EDP} are phases where semantics have been confirmed and both influence output logits, we weigh their contributions based on the confidence metric defined in Eq.~\ref{eq:confidence} when aggregating historical information. We then compute the JSD in Eq.~\ref{eq:jsd} between each pair of time steps within the historical window $W$, and use \textbf{SAP} and \textbf{EDP} as semi-U-shaped Historical Aggregation Window $\Delta_t$.

\vspace{0.1cm}
\noindent\textbf{Confidence-Weighted Aggregation.} 
While we acknowledge that the historical information within the Historical Aggregation Window is correlated with the current time step, the degree of correlation between information from different historical positions (i.e. time steps) and the current step varies.
Motivated by \citet{fu2025deepthinkconfidence}, we employ the confidence $C_i$ (for a time step $i$ within the historical window $\Delta_t$), which is formulated as: 
\vspace{-0.2cm}
\begin{equation}
C_i = -\frac{1}{\left|\Omega_t\right|} \sum_{j=1}^{\Omega_t} \log P_i(j),
\label{eq:confidence} 
\end{equation}
\noindent where $P_i(j)$ represents the probability of the $j$-th token in $\Omega_t$. The confidence $C_i$ is a token-level local confidence metric derived from the model’s internal token distribution, and high $C_i$ aligns with low entropy of the token distribution within $\Omega_t$, which denotes the degree of certainty for historical time step $i$.
Based on the mechanisms discussed above, we can obtain the historical residual signal:

\vspace{-0.2cm}
\begin{equation}
\text{logit}_\theta^\text{res}(y_t \mid T_{<{t-1}}) = \sum_{i \in \Delta_t} \left( \frac{C_i}{\sum_{j=1}^{n} C_j} \cdot \text{logit}_\theta(\hat{y}_{i} \mid T_{<{i}}) \right)
\label{eq:logit_res}
\end{equation}
\noindent where $\hat{y}_i \in \Omega_t$, and $T_{<  (\cdot )}$ represents the input to the decoder up to the time step $t-1$. For the first generated token, we use the corresponding token generated from the query $x$ as its historical information:

\vspace{-0.2cm}
\begin{equation}
T_{<i} = 
\begin{cases} 
\text{concat}(T_v, T_x, y_{<i}) & \text{if } i > 0, \\
\text{concat}(T_v, T_{x<i}) & \text{if } i \leq 0,
\end{cases}
\label{eq:tti}
\end{equation}
\noindent where $T_v$ represents the encoded visual features and $T_x$ denotes the encoded query text features. In short, for tokens generated in the early stages of generation, where sufficient complete historical information is lacking. We use the decoded tokens corresponding to $T_{x<i}$ in Eq.~\ref{eq:tti} to serve as the historical interval for these early tokens.

\subsubsection{Assembly of History and Present}
\label{sec:ahp}
While the validity of the historical residual signal has been confirmed, exclusive reliance on it in decoding often gives rise to temporal disconnection and contextual fragmentation. Thus, we need to integrate the current temporal information with the historical residual signal. This strategy can be formulated as follows:
\begin{equation}
\begin{aligned}
p_{\scriptstyle \text{ResDec}}(y_t \mid v, x, &y_{<t}) = \text{Softmax}\left[ (1-\alpha) \, \text{logit}_\theta(y_t \mid T_{<{t-1}})  \right. \\
&\quad \left. + \alpha \, \text{logit}_\theta^\text{res}(y_t \mid T_{<{t-1}}) \right]
\end{aligned}
\label{eq:res_dec} 
\end{equation}
\noindent where $\alpha \in [0,1]$ is a weight balancing the two logit streams. ($\alpha=0$ reduces to regular decoding). A larger $\alpha$ amplifies the influence of the historical residual signal $\text{logit}_\theta^\text{res}$, strengthening anchoring to context and reducing the risk of hallucination. Drawing on previous studies \cite{leng2023mitigatingobjecthallucinationslarge,li2023contrastivedecodingopenendedtext,chuang2023dola}, calibrating the entire distribution tends to penalize valid outputs of the original distribution while promoting implausible outputs derived from the augmented distribution.
\begin{equation}
\begin{aligned}
&\mathcal{V}_{head}(y_{<t}) = \left\{ y_t \in \mathcal{V}: \right. \\
& \left. p_\theta\left(y_t \mid v, x, y_{<t}\right) \geq \beta \max_{w} p_\theta\left(w \mid v, x, y_{<t}\right) \right\}
\end{aligned}
\end{equation}
\noindent where $\mathcal{V}$ is the output vocabulary of LVLMs and $\beta \in [0,1]$ is a hyperparameter for regulating the truncation of the distribution of the next token. Finally, we only consider tokens within $\mathcal{V}_{\text{head}}(y_{<t})$; for those not in $\mathcal{V}_{\text{head}}(y_{<t})$, we set their logits to $-\infty$ to effectively filter them out.

\section{Experiments}

\begin{table*}[th]
  \centering
  \setlength\tabcolsep{4pt}
  \fontsize{9}{9}\selectfont 
    \begin{tabular}{cccccccccc}
          \toprule
          \toprule
          \multirow{2.6}{*}{Models} & \multirow{2.6}{*}{Methods} & \multicolumn{2}{c}{{Random}} & \multicolumn{2}{c}{{Popular}} & \multicolumn{2}{c}{{Adversarial}} & \multicolumn{2}{c}{{Average}} \\ 
          \cmidrule(lr){3-4} \cmidrule(lr){5-6} \cmidrule(lr){7-8} \cmidrule(lr){9-10}
          & & Accuracy $\uparrow$ & F1 $\uparrow$ & Accuracy $\uparrow$ & F1 $\uparrow$ & Accuracy $\uparrow$ & F1 $\uparrow$ & Accuracy $\uparrow$ & F1 $\uparrow$ \\
          \midrule
          \multirow{10}{*}{$\text{LLaVA-1.5}$} 
          & Regular & 83.49 & 82.28 & 79.98 & 79.34 & 76.03 & 76.26 & 79.83 & 79.29 \\
          & VCD~\cite{leng2023mitigatingobjecthallucinationslarge} & 86.84 & 86.83 & 82.65 & 83.37 & 77.31 & 79.28 & 82.27 & 83.16 \\
          & ICD~\cite{wang2024mitigatinghallucinationslargevisionlanguage} & 85.11 & 84.18 & 81.08 & 80.74 & 78.16 & 78.63 & 81.45 & 81.18 \\
          & DoLa~\cite{chuang2023dola} & 84.78 & 84.19 & 79.75 & 80.61 & 76.32 & 76.16 & 80.28 & 80.32 \\
          & OPERA~\cite{huang2024operaalleviatinghallucinationmultimodal} & 87.53 & 86.45 & 84.21 & 83.50 & 80.88 & 80.69 & 84.21 & 83.55 \\

          & ONLY~\cite{wan2025onlyonelayerinterventionsufficiently} & 87.49 & 88.02 & 79.68 & 82.27 & 72.44 & 77.49 & 79.87 & 82.60 \\ %
          & AGLA~\cite{an2025mitigatingobjecthallucinationslarge} & 88.54 & 87.71 & 85.14 & 84.68 & 81.13 & 81.36 & 84.58 & 83.13 \\
          & MemVR~\cite{zou2025looktwiceanswermemoryspace} & 89.73 & 89.16 & 86.35 & 86.14 & 82.31 & \textbf{83.97} & 86.13 & 86.04 \\ %
          & VISTA~\cite{li2025hiddenlifetokensreducing} & 89.64 & 89.03 & 85.97 & 85.87 & 82.84 & 82.81 & 86.15 & 86.29 \\ %
          & \textit{ResDec} (ours) & \textbf{90.21} & \textbf{89.87} & \textbf{86.97} & \textbf{87.22} & \textbf{84.50} & 83.69 & \textbf{87.23} & \textbf{86.93} \\
          
          \midrule
          \multirow{9}{*}{$\text{InstructBLIP}$} 
          & Regular & 80.42 & 80.94 & 76.09 & 77.65 & 72.37 & 75.42 & 76.29 & 78.00 \\
          & VCD~\cite{leng2023mitigatingobjecthallucinationslarge} & 84.11 & 84.13 & 79.94 & 80.80 & 76.32 & 78.08 & 80.12 & 81.00 \\
          & ICD~\cite{wang2024mitigatinghallucinationslargevisionlanguage} & 82.37 & 83.32 & 78.13 & 79.12 & 75.12 & 76.24 & 78.54 & 79.56 \\
          & DoLa~\cite{chuang2023dola} & 83.00 & 83.00 & 78.99 & 79.85 & 74.67 & 76.68 & 78.89 & 79.84 \\
          & OPERA~\cite{huang2024operaalleviatinghallucinationmultimodal} & 85.07 & 84.39 & 78.33 & 80.20 & 75.50 & 78.17 & 79.63 & 80.92 \\
          
          & ONLY~\cite{wan2025onlyonelayerinterventionsufficiently} & 87.18 & 88.15 & 80.43 & 82.07 & 74.83 & 78.00 & 80.81 & 82.74 \\ %
          & AGLA~\cite{an2025mitigatingobjecthallucinationslarge} & 87.30 & 87.07 & 81.86 & 82.58 & 77.29 & 79.16 &  82.15 & 82.94 \\
          & VISTA~\cite{li2025hiddenlifetokensreducing} & 87.45 & 87.66 & 81.34 & 80.94 & \textbf{80.82} & 79.67 & 83.20 & 82.76 \\ %
          & \textit{ResDec} (ours) & \textbf{88.32} & \textbf{89.11} & \textbf{82.01} & \textbf{83.42} & 80.41 & \textbf{81.87} & \textbf{83.58} & \textbf{84.80} \\ %

          \midrule  
          \multirow{8}{*}{$\text{Qwen2.5-VL}$} 
          & Regular & 87.39 & 86.31 & 86.03 & 84.56 & 84.90 & 83.35 & 86.11 & 84.74 \\ %
          & VCD~\cite{leng2023mitigatingobjecthallucinationslarge} & 88.67 & 88.21 & 87.84 & 85.78 & 85.45 & 85.97 & 87.32 & 86.65 \\ %
          & ICD~\cite{wang2024mitigatinghallucinationslargevisionlanguage} & 88.67 & 88.21 & 87.84 & 85.78 & 85.45 & 85.97 & 87.32 & 86.65 \\ %
          & DoLa~\cite{chuang2023dola} & 87.98 & 87.71 & 87.01 & 85.28 & 85.67 & 84.88 & 86.89 & 85.96 \\ %
          & OPERA~\cite{huang2024operaalleviatinghallucinationmultimodal} & 89.97 & 90.32 & 88.89 & 87.05 & 86.79 & 87.35 & 88.55 & 88.24 \\ %

          & ONLY~\cite{wan2025onlyonelayerinterventionsufficiently} & 90.94 & 90.62 & 89.23 & 88.21 & 87.32 & 86.63 & 89.16 & 88.49 \\ %
          & VISTA~\cite{li2025hiddenlifetokensreducing} & 90.53 & 90.77 & 87.92 & 88.95 & 88.03 & \textbf{87.24} & 88.83 & 88.99 \\ %
          & \textit{ResDec} (ours) & \textbf{91.17} & \textbf{91.62} & \textbf{90.34} & \textbf{89.83} & \textbf{88.96} & \textbf{87.24} & \textbf{90.16} & \textbf{89.56} \\ %
          
          \bottomrule
          \bottomrule
      \end{tabular}
    \caption{\label{tab:POPE}
      Experimental results on the three POPE subsets with LLaVA-1.5, InstructBLIP and Qwen2.5-VL. The best performances in each setting are highlighted in \textbf{bold}.
    }
    \vspace{-0.2cm}
    
\end{table*}

\label{sec:experiments}
\subsection{Experiment Setup}
\noindent \textbf{Benchmarks.} 
To rigorously assess the effectiveness of our \textit{ResDec}, we conduct extensive experiments on eleven benchmarks: POPE~\cite{li-etal-2023-evaluating}, CHAIR~\cite{rohrbach-etal-2018-object}, 
HallusionBench~\cite{guan2024hallusionbench}, 
MME~\cite{fu2025mmecomprehensiveevaluationbenchmark}, MMBench~\cite{liu2024mmbenchmultimodalmodelallaround}, MMVP~\cite{tong2024eyeswideshutexploring}, ScienceQA~\cite{lu2022learnexplainmultimodalreasoning}, MMStar~\cite{chen2024rightwayevaluatinglarge}, MM-Vet~\cite{yu2024mmvetevaluatinglargemultimodal}, $\text{SEEDBench2}\_\text{Plus}$~\cite{li2024seedbench2plusbenchmarkingmultimodallarge}, LLaVA-Bench (In-the-Wild)~\cite{NEURIPS2023_6dcf277e}. More details can be found in Appendix~\ref{appendix:datasets}.

\noindent \textbf{Evaluated LVLMs and Baselines.} We assess the effectiveness of our \textit{ResDec} strategy on three contemporary LVLMs: LLaVA-1.5 (7B)~\cite{Liu_2024_CVPR}, Qwen2.5-VL-Instruct (7B)~\cite{bai2025qwen25vltechnicalreport}, and InstructBLIP (Vicuna-7B)~\cite{dai2023instructblipgeneralpurposevisionlanguagemodels}, and compare its performance with several state-of-the-art methods, including VCD~\cite{leng2023mitigatingobjecthallucinationslarge}, ICD~\cite{wang2024mitigatinghallucinationslargevisionlanguage}, DoLa~\cite{chuang2023dola}, OPERA~\cite{huang2024operaalleviatinghallucinationmultimodal}, AGLA~\cite{an2025mitigatingobjecthallucinationslarge}, ONLY~\cite{wan2025onlyonelayerinterventionsufficiently}, MemVR~\cite{zou2025looktwiceanswermemoryspace}, and VISTA~\cite{li2025hiddenlifetokensreducing}. Throughout our experiments, we set $\alpha = 0.5$ and $\beta = 0.1$ for \textit{ResDec}, unless explicitly stated otherwise. More implementation details are provided in Appendix~\ref{appendix:implementation_details}.

\subsection{Performance on Hallucination Benchmarks}
\begin{table}[th]
  \centering
  \setlength\tabcolsep{4pt}
  \fontsize{7}{6.5}\selectfont 
  \resizebox{\linewidth}{!}{
      \begin{tabular}{c|c|ccccc}
            \toprule
            \toprule
            \multirow{2.6}{*}{Models} & \multirow{2.6}{*}{Methods} & 
            \multicolumn{3}{c}{HallusionBench} & \multicolumn{2}{c}{CHAIR}  \\
            \cmidrule(lr){3-5} \cmidrule(lr){6-7}
            & & fACC $\uparrow$ & qACC $\uparrow$ & aACC $\uparrow$ & $\text{CHAIR}_S$ $\downarrow$ & $\text{CHAIR}_I$ $\downarrow$ \\
            \midrule
            \multirow{9}{*}{$\text{LLaVA-1.5}$} 
            & Regular & 17.9 & 8.1 & 41.5 & 55.0 & 16.3 \\
            & OPERA~\cite{huang2024operaalleviatinghallucinationmultimodal} & 16.2 & 5.5 & 41.2 & 47.8 & 14.6 \\
            & ICD~\cite{wang2024mitigatinghallucinationslargevisionlanguage} & 13.9 & 8.4 & 38.2 & 56.2 & 16.3\\
            & VCD~\cite{leng2023mitigatingobjecthallucinationslarge} & 13.9 & 11.4 & 41.1 & 54.4 & 16.6 \\
            & DoLa~\cite{chuang2023dola} & 15.7 & 9.3 & 41.5 & 57.0 & 15.9 \\
            & AGLA~\cite{an2025mitigatingobjecthallucinationslarge} & 17.2 & 8.3 & 42.9 & 43.3 & 14.1  \\
            & ONLY~\cite{wan2025onlyonelayerinterventionsufficiently} & 17.5 & 9.3 & 42.7 & 49.8 & 14.3 \\
            & MemVR~\cite{zou2025looktwiceanswermemoryspace} & 17.9 & 9.0 & 42.5 & 46.6 & 13.0 \\
            & \textit{ResDec} (ours) & \textbf{18.2} & \textbf{12.7} & \textbf{45.1} & \textbf{42.7} & \textbf{12.6} \\

            \midrule 
            \multirow{7}{*}{$\text{InstructBLIP}$}
            & Regular & 10.1 & 9.5 & 45.3 & 54.0 & 18.1 \\
            & OPERA~\cite{huang2024operaalleviatinghallucinationmultimodal} & 9.7 & 8.2 & 42.3 & 53.7 & 12.8  \\
            & ICD~\cite{wang2024mitigatinghallucinationslargevisionlanguage} & 7.2 & 7.5 & 41.0 & 59.3 & 17.5 \\
            & VCD~\cite{leng2023mitigatingobjecthallucinationslarge} & 8.1 & 8.5 & 43.3 & 57.0 & 17.0  \\
            & DoLa~\cite{chuang2023dola} & 8.4 & 8.6 & 43.8 & 60.0 & 20.1 \\
            & AGLA~\cite{an2025mitigatingobjecthallucinationslarge} & 9.7 & 9.2 & 45.9 & 49.0 & 12.1  \\
            & \textit{ResDec} (ours) & \textbf{11.3} & \textbf{10.2} & \textbf{46.5} & \textbf{47.7} & \textbf{11.3} \\
            
            \midrule 
            \multirow{6}{*}{$\text{Qwen2.5-VL}$}
            & Regular & 43.4 & 45.8 & 69.1 & 30.6 & 8.4 \\
            & OPERA~\cite{huang2024operaalleviatinghallucinationmultimodal} & 41.5 & 43.4 & 68.7 & 28.4 & 7.6  \\
            & ICD~\cite{wang2024mitigatinghallucinationslargevisionlanguage} & 38.7 & 39.8 & 67.6 & 30.9 & 8.3 \\
            & VCD~\cite{leng2023mitigatingobjecthallucinationslarge} & 39.8 & 41.3 & 68.7 & 27.9 & 7.8  \\
            & DoLa~\cite{chuang2023dola} & 40.8 & 39.5 & 66.9 & 28.2 & 8.9  \\
            & \textit{ResDec} (ours) & \textbf{47.1} & \textbf{49.0} & \textbf{71.6} & \textbf{25.8} & \textbf{6.8} \\

            \bottomrule
            \bottomrule
        \end{tabular}
        }
      \caption{\label{tab:hallusionbench}
        HallusionBench and CHAIR evaluation results of different methods. We limit the maximum number of new tokens $128$. Lower ($\downarrow$) $\text{CHAIR}_S$ and $\text{CHAIR}_I$ denote better performance. The best performances in each setting are highlighted in \textbf{bold}.
      }
    \vspace{-0.5cm}
\end{table}

\begin{table*}[th]
  \centering
  \setlength\tabcolsep{3pt}
  \fontsize{8.4}{8}\selectfont 
    
    \begin{tabular}{cccccccccccc}
          \toprule
          \toprule
          Models & Methods & MME $\uparrow$ & MMBench $\uparrow$ & 
          ScienceQA $\uparrow$ & 
          MMVP $\uparrow$ & MMStar $\uparrow$ & MM-Vet $\uparrow$ & $\text{SEEDBench2}\_\text{Plus}$ $\uparrow$ & LLaVA-Bench $\uparrow$\\
          \midrule
          \multirow{9}{*}{$\text{LLaVA-1.5}$} 
          & Regular & 1810.41 & 64.24 & 68.02 & 20.67 & 32.67 & 31.10 & 38.65 & 61.80 \\
          & VCD~\cite{leng2023mitigatingobjecthallucinationslarge} & 1818.67 & 55.14 & 62.96  & 20.33 & 29.20 & 30.20 & 39.31 & 58.50 \\
          & ICD~\cite{wang2024mitigatinghallucinationslargevisionlanguage} & 1757.28 & 39.03 & 51.61 & 17.00 & 25.80 & 26.30 & 35.40 & 55.10 \\
          & DoLa~\cite{chuang2023dola} & 1829.72 & 63.78 & 68.32 & 20.67 & 31.67 & 30.70 & 38.21 & 60.70 \\
          & OPERA~\cite{huang2024operaalleviatinghallucinationmultimodal} & 1844.51 & 64.32 & 68.77 & 21.00 & 33.00 & 32.00 & 39.75 & 61.40 \\

          & ONLY~\cite{wan2025onlyonelayerinterventionsufficiently} & 1876.35 & 64.87 & 67.23 & 21.33 & 32.27 & 32.50 & 37.90 & 61.60 \\ %
          & MemVR~\cite{zou2025looktwiceanswermemoryspace} & 1864.68 & 65.12 & 68.96 & 23.67 & 32.73 & 32.80 & 40.84 & 62.20 \\ %
          & VISTA~\cite{li2025hiddenlifetokensreducing} & 1857.64 & 65.00 & 68.22 & 20.33 & 32.60 & 32.00 & 40.49 & 60.90 \\ %
          & \textit{ResDec} (ours) & \textbf{1881.57} & \textbf{65.70} & \textbf{69.46} & \textbf{24.00} & \textbf{33.40} & \textbf{33.20} & \textbf{41.15} & \textbf{62.60} \\
          
          \midrule  
          \multirow{7}{*}{$\text{InstructBLIP}$} 
          & Regular & 1385.25 & 38.32 & 60.49 & 17.33 & 32.20 & 33.10 & 30.87 & 59.80 \\ %
          & VCD~\cite{leng2023mitigatingobjecthallucinationslarge} & 1408.15 & 29.13 & 56.42 & 18.00 & 31.67 & 29.30 & 28.55 & 61.10 \\ %
          & ICD~\cite{wang2024mitigatinghallucinationslargevisionlanguage} & 1375.48 & 21.28 & 49.78 & 15.00 & 24.20 & 23.60 & 26.39 & 51.00 \\ %
          & DoLa~\cite{chuang2023dola} & 1398.95 & 38.81 & 58.21 & 17.33 & 31.27 & 32.50 & 30.61 & 56.60  \\ %
          & OPERA~\cite{huang2024operaalleviatinghallucinationmultimodal} & 1406.54 & 39.31 & 60.83 & 17.67 & 30.67 & 33.70 & 31.80 & 63.10 \\ %

          & VISTA~\cite{li2025hiddenlifetokensreducing} & 1413.05 & 39.73 & 60.93 & 18.33 & 32.40 & 33.80 & 32.02 & 63.50 \\ %
          & \textit{ResDec} (ours) & \textbf{1426.83} & \textbf{40.13} & \textbf{62.92} & \textbf{19.00} & \textbf{32.67} & \textbf{34.30} & \textbf{33.38} & \textbf{64.40} \\ %

          \midrule  
          \multirow{7}{*}{$\text{Qwen2.5-VL}$} 
          & Regular & 2309.42 & 81.05 & 88.99 & 58.00 & 63.87 & 67.10 & 69.61 & 91.00 \\ %
          & VCD~\cite{leng2023mitigatingobjecthallucinationslarge} & 2296.78 & 69.23 & 90.08 & 59.67 & 63.60 & 64.70 & 67.90 & 89.50  \\ %
          & ICD~\cite{wang2024mitigatinghallucinationslargevisionlanguage} & 2284.41 & 60.06 & 84.83 & 55.33 & 57.20 & 57.70 & 62.32 & 83.70  \\ %
          & DoLa~\cite{chuang2023dola} & 2313.23 & 79.03 & 88.75  & 59.33 & 64.20 & 65.60 & 67.19 & 87.30  \\ %
          & OPERA~\cite{huang2024operaalleviatinghallucinationmultimodal} & 2318.13 & 81.87 & 90.33 & 61.67 & 64.53 & 66.90 & 68.99 & 91.30  \\ %

          & VISTA~\cite{li2025hiddenlifetokensreducing} & 2324.93 & 81.04 & 90.23 & 60.33 & 64.27 & 67.80 & 69.30 & 91.70 \\ %
          & \textit{ResDec} (ours) & \textbf{2348.40} & \textbf{82.64} & \textbf{90.48} & \textbf{63.33} & \textbf{65.47} & \textbf{68.70} & \textbf{70.31} & \textbf{91.90} \\ %
          
          \bottomrule
          \bottomrule
      \end{tabular}
    \caption{\label{tab:General}
      Experimental results on the eight comprehensive benchmarks with LLaVA-1.5, InstructBLIP and Qwen2.5-VL. The best performances in each setting are highlighted in \textbf{bold}.
    }
    \vspace{-0.2cm}
\end{table*}

\noindent \textbf{Experiments on POPE.}
Table~\ref{tab:POPE} shows experimental results on the three POPE dataset with LLaVA-1.5, InstructBLIP and Qwen2.5-VL. \textit{ResDec} consistently outperforms the ``Regular'' decoding strategy by substantial margins (average improvements of $7.84\%$ in accuracy and $8.01\%$ in F1 score) across three LVLMs and settings. Additionally, \textit{ResDec} clearly outperforms all compared SOTA methods, demonstrating its effectiveness in hallucination mitigation. 

\noindent \textbf{Experiments on HallusionBench and CHAIR.}
As shown in Table~\ref{tab:hallusionbench}, \textit{ResDec} surpasses regular decoding and other methods across all models on HallusionBench, delivering state-of-the-art performance with consistent gains across metrics. We further assess open-ended captioning on CHAIR; relative to regular decoding, \textit{ResDec} yields average improvements of $16.57\%$ on $\text{CHAIR}_S$ and $26.44\%$ on $\text{CHAIR}_I$ with significant enhancements across InstructBLIP, LLaVA-1.5 and Qwen2.5-VL. Collectively, these results indicate that \textit{ResDec} effectively mitigates hallucinations while enhancing open-ended caption generation, highlighting its versatility and robust overall performance.

\subsection{Performance on Comprehensive Benchmarks} 
We evaluate the performance of \textit{ResDec} on comprehensive benchmarks, including MM-Vet, MME, MMBench, MMVP, MMStar, and $\text{SEEDBench2}\_\text{Plus}$. As shown in Table~\ref{tab:General}, \textit{ResDec} consistently outperforms competing methods across LLaVA-1.5, InstructBLIP, and Qwen2.5-VL, demonstrating its robust generalizability. For example, \textit{ResDec} outperforms ``Regular'' decoding by $2.87\%$, $2.99\%$, and $11.65\%$ on MME, ScienceQA, and MMVP, respectively, across three models.  
This consistency across different LVLMs and benchmarks highlights the effectiveness of \textit{ResDec} in mitigating hallucinations and enhancing multimodal reasoning, aligning visual and linguistic information at both object and attribute levels. 
\subsection{Ablation Study}

\begin{table}[t]
  \centering
  \setlength\tabcolsep{3pt}
  \fontsize{8}{8}\selectfont 
      \begin{tabular}{cc|cccc}
            \toprule
            \multirow{2.6}{*}{$\alpha$} & \multirow{2.6}{*}{$\beta$} & MME & \multicolumn{2}{c}{POPE} & MMStar \\
            \cmidrule(lr){3-3} \cmidrule(lr){4-5} \cmidrule(lr){6-6}
            & & Accuracy $\uparrow$ & Accuracy $\uparrow$ & F1 $\uparrow$ & Accuracy $\uparrow$ \\ 
            
            \midrule
            0.25 & 0.1 & 2326.31 & 89.64 & 89.11 & 64.20 \\
            0.5 & 0.1 & \textbf{2348.40} & \textbf{90.16} & \textbf{89.56} & \textbf{65.40} \\
            0.75 & 0.1 & 1875.31 & 82.56 & 81.93 &  62.67 \\
            1.0 & 0.1 & 1583.17 & 72.50 & 71.73 & 61.80 \\
            \midrule
            
            0.5 & 0 & 2221.73 & 87.21 & 87.73 & 64.20 \\
            0.5 & 0.001 & 2184.91 & 88.02 & 87.21 & 63.47 \\
            0.5 & 0.01 & 2274.15 & 88.74 & 88.27 & 64.80 \\
            0.5 & 0.2 & 2348.40 & \textbf{90.35} & \textbf{89.96} & 65.67 \\
            0.5 & 0.5 & \textbf{2353.21} & 89.83 & 89.51 & \textbf{66.07} \\
            \bottomrule
        \end{tabular}
      \caption{\label{tab:Hyperparameters}
        An ablation study of $\alpha$ and $\beta$ in Qwen2.5-VL on MME, POPE and MMStar benchmarks. The bset performance in each setting is emphasized in \textbf{boldface}.
      }
      \vspace{-0.6cm}
\end{table}

\begin{figure}[t]
  \centering
   \includegraphics[width=1.0\linewidth]{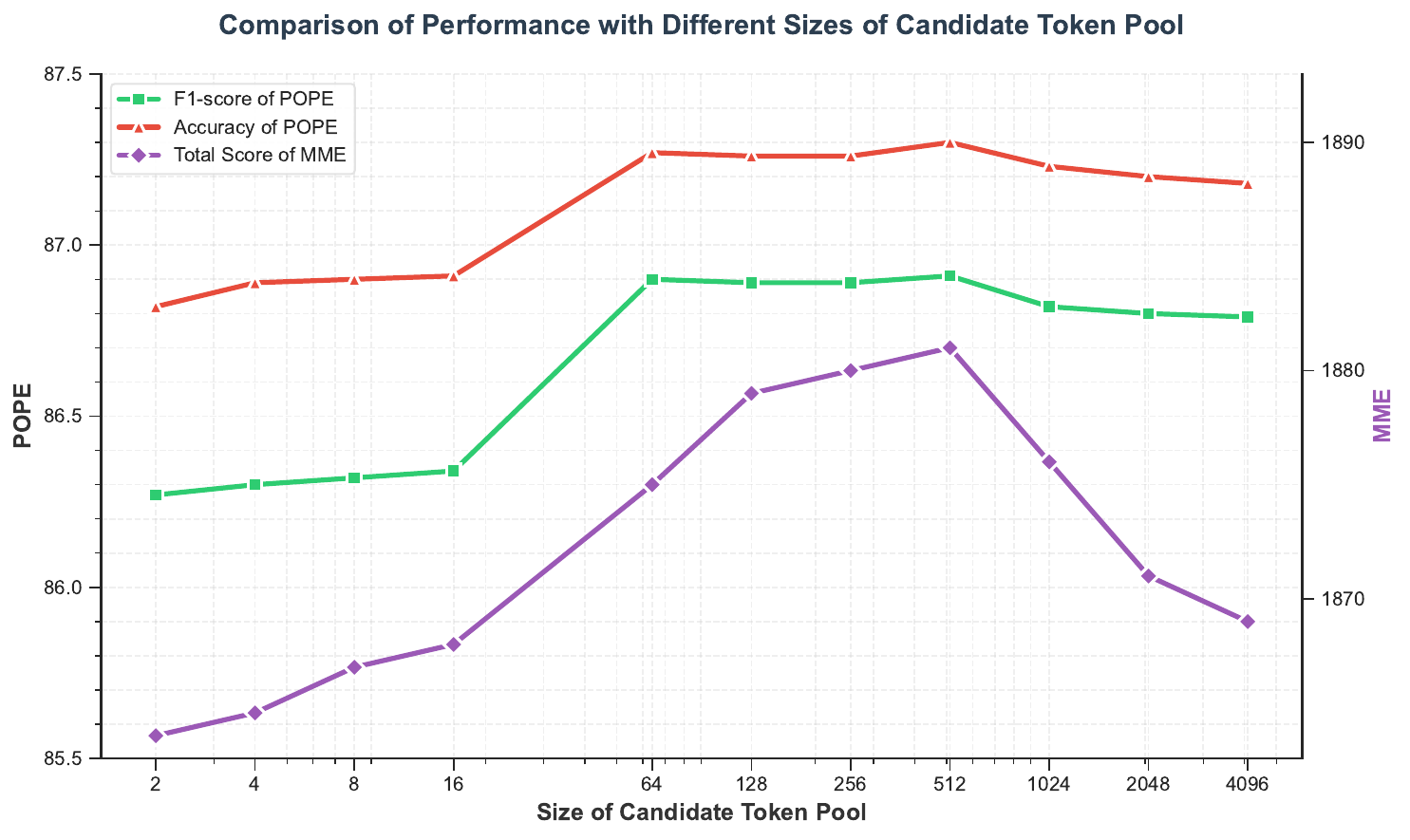}

   \caption{Ablation study on different sizes of Candidate Token Pool conducted on LLaVA-1.5 and focusing on POPE and MME. The optimal size of Candidate Token Pool ranges from $64$ to $512$, while excessively small or large sizes degrade performance by impairing \textbf{SAP} localization.}
   \label{fig:token_pool}
   \vspace{-0.4cm}
\end{figure}

\noindent \textbf{Size of Candidate Token Pool.} In the paragraph, we explore the size of Candidate Token Pool $\left| \Omega_t \right|$ to observe its impact on \textit{ResDec}. To determine the optimal size of the candidate token pool, we explore sizes ranging from $2$ to $4096$. The results shown in Fig.~\ref{fig:token_pool} indicate that the most suitable size for the Candidate Token Pool, denoted as $\Omega_t$, falls between $64$ and $512$ tokens. Both excessively small and large sizes lead to inaccurate localization of \textbf{SAP}, which in turn causes performance degradation. This trend can be explained as follows: When $\Omega_t$ is too small, the pool fails to capture comprehensive variations in JSD, making it insufficient to reflect the critical differences required for accurate \textbf{SAP} localization. Conversely, an excessively large $\Omega_t$ inadvertently incorporates too much local information from earlier time steps, which distorts the JSD calculation, rendering the divergence metric inaccurate and ultimately undermining the precision of \textbf{SAP} localization.

\begin{figure*}[th]
  \centering
   \includegraphics[width=1.0\linewidth]{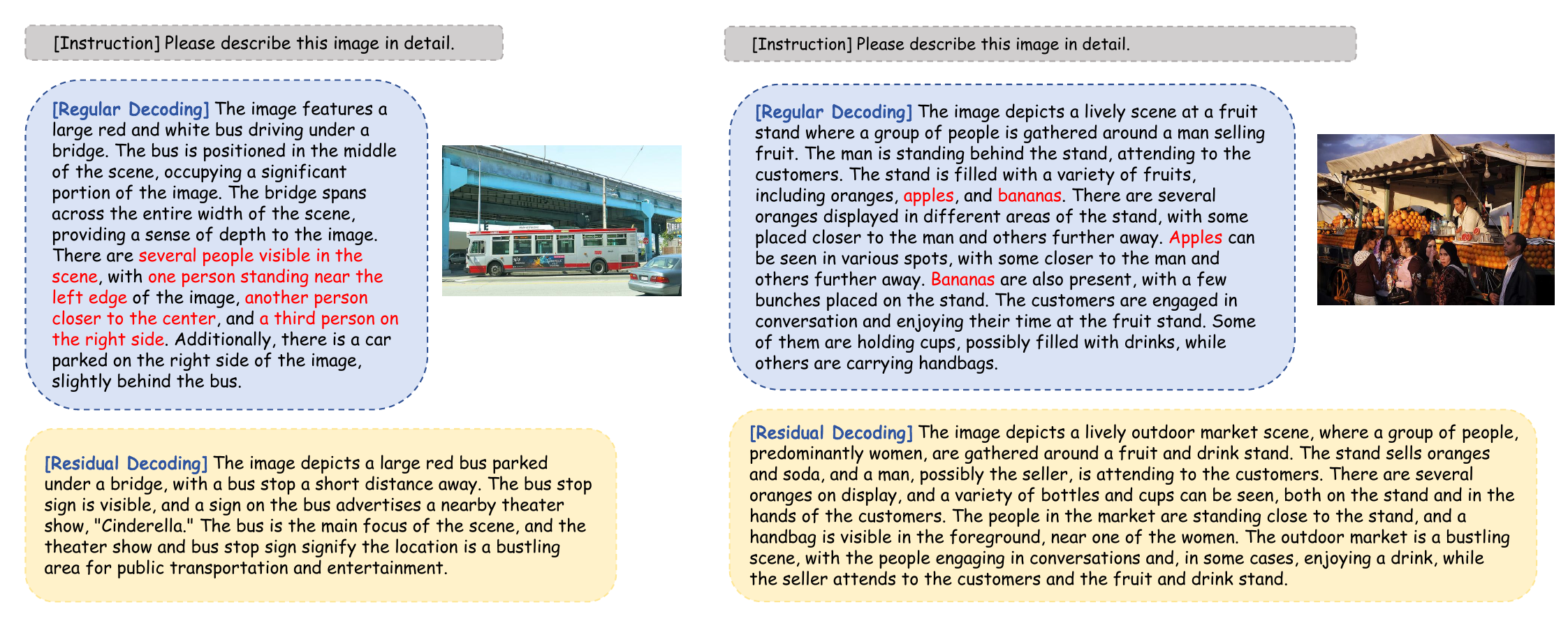}

   \caption{Case study comparing regular decoding and residual decoding in image captioning. Hallucinated details in ``regular'' decoding outputs are highlighted in \textcolor{red}{red}.}
   \label{fig:case}
   \vspace{-0.2cm}
   
\end{figure*} 
\noindent \textbf{Effect of Hyperparameters in \textit{ResDec}.} To investigate the impact of hyperparameters $\alpha$ and $\beta$ on \textit{ResDec}, we conduct an ablation study on three benchmarks. $\alpha$ modulates the amplification between the historical residual signal and the original output distributions in Eq.~\ref{eq:res_dec}. The historical signal helps rectify hallucination issues by retaining valid prior information.  
As shown in Table~\ref{tab:Hyperparameters}, $\alpha$ performs well below 0.5, but higher values ($\alpha = 0.75, 1.0$) often cause adverse effects, as the historical signal is an auxiliary correction, not directly usable for decoding.  
Meanwhile, $\beta$ controls the adaptive plausible constraint in \textit{ResDec}, with larger values indicating more aggressive truncation. Setting $\beta=0$ leads to suboptimal performance, while $\beta > 0$ improves performance, validating the constraint’s effectiveness.

\subsection{Analysis}
\noindent \textbf{Effect of Different Decoding Strategies.} Besides direct sampling, we also present an ablation study with various decoding strategies on the POPE and MME datasets using LLaVA-1.5. The experiment includes three additional decoding strategies: Nucleus Sampling ($p$ = 0.7), Top-K sampling ($k$ = 50), Temperature Sampling ($t$ = 0.5), and Greedy decoding. The results in Table~\ref{tab:decoding_strategies} demonstrate that our \textit{ResDec} consistently mitigates hallucinations, regardless of the adopted decoding strategy. This highlights the robustness of \textit{ResDec} in improving text generation quality and accuracy across different decoding strategies.

\begin{table}[t]
  \centering
  \setlength\tabcolsep{5pt}
  \fontsize{10}{10}\selectfont 
  \resizebox{\linewidth}{!}{
  
      \begin{tabular}{cc|ccc}
            \toprule
            
            \multicolumn{2}{c|}{\multirow{2.5}{*}{Decoding Strategy}} 
            & \multicolumn{2}{c}{POPE} & MME \\
            \cmidrule(lr){3-4} \cmidrule(lr){5-5}
            
            & & Accuracy $\uparrow$ & F1 $\uparrow$ & Accuracy $\uparrow$ \\
            \midrule
            \multirow{2}{*}{Nucleus Sampling}  
            & Vanilla & 77.61 & 78.43 & 1792.35  \\
            & \textit{ResDec} & \textbf{84.15} & \textbf{85.82} & \textbf{1846.82}  \\
            \midrule
            
            \multirow{2}{*}{Top-K Sampling} 
            & Vanilla & 76.62 & 77.32 & 1773.67  \\
            & \textit{ResDec} & \textbf{83.82} & \textbf{84.05} & \textbf{1852.09} \\
            \midrule
            
            \multirow{2}{*}{Temperature Sampling} 
            & Vanilla & 79.13 & 78.56 & 1803.16  \\
            & \textit{ResDec} & \textbf{86.62} & \textbf{85.82} & \textbf{1872.91}  \\
            \midrule
            
            \multirow{2}{*}{Greedy} 
            & Vanilla & 79.83 & 79.29 & 1810.41  \\
            & \textit{ResDec} & \textbf{87.23} & \textbf{86.93} & \textbf{1881.57}  \\

            \bottomrule
        \end{tabular}
    }
      \caption{\label{tab:decoding_strategies} 
      An ablation study of different decoding strategies. ``Vanilla'' means sampling from the original distribution.
      }
      \vspace{-0.4cm}
      
\end{table}

\noindent \textbf{Efficiency Comparison.}
To analyze the performance of \textit{ResDec} in terms of efficiency metrics, we refer to Table~\ref{tab:Efficiency}. \textit{ResDec} needs to maintain a sequence of historical logits during operation. Crucially, by reusing historical logits naturally generated during inference, \textit{ResDec} incurs negligible additional latency and memory overhead. Compared with existing SOTA methods for hallucination mitigation, our \textit{ResDec} achieves the lowest latency (only $0.02 \times$ slower than Greedy decoding). These results highlight the practicality of \textit{ResDec}, which achieves an optimal balance between hallucination mitigation and computational efficiency.

\begin{table}[th]
  \centering
  \setlength\tabcolsep{5pt}
  \fontsize{10}{10}\selectfont 
  \resizebox{\linewidth}{!}{
  
      \begin{tabular}{c|ccc}
            \toprule
            \multirow{2}{*}{Methods} 
            & Latency $\downarrow$ & Throughput $\uparrow$ & Memory Cost $\downarrow$ \\
            & (ms/token) & (token/s) & (MB) \\
            
            \midrule
            Greedy & 28.54 & 35.04 & 14257 \\
            OPERA~\cite{huang2024operaalleviatinghallucinationmultimodal} & 104.46 & 9.57 & 21300 \\
            ICD~\cite{wang2024mitigatinghallucinationslargevisionlanguage} & 30.82 & 32.44 & \textbf{14263} \\
            VCD~\cite{leng2023mitigatingobjecthallucinationslarge} & 62.79 & 15.93 & 14967 \\
            ONLY~\cite{wan2025onlyonelayerinterventionsufficiently} & 30.54 & 32.75 & 14951 \\
            MemVR~\cite{zou2025looktwiceanswermemoryspace} & \underline{29.68} & \underline{33.69} & 14345 \\
            VISTA~\cite{li2025hiddenlifetokensreducing} & 36.32 & 27.53 & 14783 \\
            \textit{ResDec} (ours) & \textbf{29.11} & \textbf{34.35} & \underline{14296} \\
            
            \bottomrule
        \end{tabular}
    }
      \caption{\label{tab:Efficiency}
        Performance comparison of SOTA methods and \textit{ResDec} in latency, throughput, and memory usage. The best and suboptimal results are highlighted in \textbf{bold} and \underline{underlined}, respectively.
      }
      \vspace{-0.4cm}
\end{table}
\noindent \textbf{Case Study.} Fig.~\ref{fig:case} presents two case studies showing how, under the same visual input and prompt, regular decoding tends to introduce hallucinated details not visible in the images, such as \textit{additional people} and \textit{specific fruit types}, based on language priors and pretraining co-occurrence patterns. In contrast, residual decoding reduces these hallucinations while maintaining coherence and visual alignment. More examples and ablation studies are in Appendix~\ref{appendix:case_study}.
\section{Conclusion}
In this paper, we investigated the root cause of hallucinations in LVLMs and revealed that hallucinated tokens emerge and surpass visually grounded tokens during decoding. Based on this, we introduced Residual Decoding (\textit{ResDec}), a training-free decoding strategy that mitigates hallucinations with negligible inference overhead. \textit{ResDec} leverages the semantic clarity of historical token distributions and integrates residual guidance into decoding, effectively suppressing Language-Prior Hallucination. Extensive experiments on hallucination benchmarks show \textit{ResDec} alleviates hallucinations while preserving descriptive coherence and efficiency. This work offers insights for decoding-stage interventions in LVLMs and paves the way for more reliable, visually grounded generation.
{
    \small
    \bibliographystyle{ieeenat_fullname}
    \bibliography{main}
}

\clearpage
\setcounter{page}{1}
\maketitlesupplementary

\appendix
\section{Implementation Details}
\label{appendix:implementation_details}

\subsection{Datasets}
\label{appendix:datasets}

\paragraph{MME~\cite{fu2025mmecomprehensiveevaluationbenchmark}} evaluates the capabilities of MLLMs, dividing the evaluation into two major
categories: perception and cognition. The perception category includes fine-grained tasks such as existence, count, location,
rough color, poster, celebrity, scene, landmark, artwork identification, and OCR. The cognition category includes tasks like
commonsense reasoning, numerical calculations, text translation, and code reasoning. All questions in this benchmark are
structured to be answered with a simple yes or no.

\paragraph{POPE~\cite{li2023evaluatingobjecthallucinationlarge}} is a metric based on Visual Question Answering (VQA) designed to evaluate hallucinations in MLLMs. This metric assesses the model's response to the question, ``Is [object] in this image?'' To clarify that this is a binary VQA task, the prompt is supplemented with the instruction, ``Please answer yes or no.'' POPE is built on three core data sources: COCO, A-OKVQA, and GQA. We report the overall accuracy and F1-score.

\paragraph{CHAIR~\cite{rohrbach-etal-2018-object}} is used to assess the alignment between generated captions and the content of the given image. The CHAIR metric consists of two evaluation types: per-instance evaluation ($\text{CHAIR}_I$) and per-sentence evaluation ($\text{CHAIR}_S$), which are defined as follows:
\begin{equation}
\text{CHAIR}_i = \frac{|\{\text{hallucinated objects}\}|}{|\{\text{all objects mentioned}\}|}
\end{equation}

\begin{equation}
\text{CHAIR}_s = \frac{|\{\text{sentences with hallucinated object}\}|}{|\{\text{ all sentences}\}|}
\end{equation}

\paragraph{HallusionBench~\cite{guan2024hallusionbench}.}
HallusionBench is a benchmark constructed to evaluate hallucination robustness of large vision–language models under challenging, misleading scenarios. It contains 951 carefully curated question–answer pairs, where each question is designed to probe whether a model can resist visually or semantically deceptive cues and avoid generating hallucinated content. This benchmark thus provides a focused testbed for analyzing and improving the reliability of multimodal reasoning.

\paragraph{MMBench~\cite{liu2024mmbenchmultimodalmodelallaround}}is a comprehensive benchmark for evaluating general purpose vision language models. It contains carefully curated multiple choice questions in both English and Chinese that test perception, reasoning, and understanding across diverse real world images and scenarios.

\paragraph{ScienceQA~\cite{lu2022learnexplainmultimodalreasoning}} is a multimodal multiple-choice benchmark that evaluates scientific reasoning over grade-school curricula, with questions accompanied by textual context and often images or diagrams. In our experiments, we adopt the official ScienceQA test split, which consists of 2,017 question–answer pairs, and use it to assess the model’s ability to perform multimodal reasoning and science knowledge understanding.

\paragraph{MMStar~\cite{chen2024rightwayevaluatinglarge}} is a comprehensive benchmark for evaluating large vision–language models, consisting of 1,500 carefully curated multiple-choice questions that cover a wide range of real-world images and tasks. It is designed to probe fine-grained multimodal abilities, including perception, reasoning, and instruction following, thereby enabling more reliable and nuanced assessment of general-purpose multimodal models.

\paragraph{$\text{SEEDBench2}\_\text{Plus}$~\cite{li2024seedbench2plusbenchmarkingmultimodallarge}}
 is a large scale benchmark for evaluating multimodal large language models. It contains 2,277 question–answer pairs covering diverse perception and reasoning skills, and provides a fine grained assessment of both visual understanding and high level multimodal reasoning.

\paragraph{MM-Vet~\cite{yu2024mmvetevaluatinglargemultimodal}}
is a benchmark designed to assess large vision–language models on challenging, multi-step tasks. It specifies six core capabilities in the vision–language setting: recognition, optical character recognition, factual and commonsense knowledge, language generation, spatial understanding, and mathematical reasoning. To obtain consistent and reliable scores across heterogeneous question formats, it employs a language model as an automatic evaluator. The benchmark contains 187 images collected from diverse online sources and 205 questions, each constructed so that answering it requires one or more of the defined capabilities.

\paragraph{MMVP.} MMVP~\cite{tong2024eyeswideshutexploring} is a benchmark targeting the fine-grained visual recognition ability of LVLMs based on CLIP-blind image pairs. It contains 150 pairs of images, with each pair accompanied by a binary-choice question. The two images in a pair are evaluated independently, and an LVLM receives credit for a pair only if it answers the questions for both images correctly.

\paragraph{LLaVA-Bench.}
LLaVA-Bench (In-the-Wild)~\cite{NEURIPS2023_6dcf277e} consists of 24 images representing various complex scenes, memes, paintings, and sketches, paired with 60 challenging questions. A subset of this dataset is used for qualitative comparisons of responses generated by different decoding strategies. we assess the accuracy and depth of the generated responses using GPT-4V.

\section{Theoretical Analysis of ResDec}
\subsection{Derivation of Language Priors}
\label{appendix:language_priors}

Following \citet{lin2024revisitingrolelanguagepriors}, we define the \emph{language prior} as the text-only, vision-agnostic conditional distribution $P(y \mid x)$, which represents the generative tendency of the output sequence $y = (y_1, \dots, y_T)$ given only the textual context $x$, without considering the visual input $v$. This language prior is related to the joint distribution of text and vision via the equivalent factorizations:

{
\small
\begin{equation}
P(y, v \mid x) = P(y \mid x)\,P(v \mid y, x) = P(v \mid x)\,P(y \mid v, x),
\end{equation}
}

\noindent where $P(y \mid x)$ isolates the language-driven preference over the outputs once the visual evidence $v$ is removed. These factorizations highlight how the text-only distribution and visual context interact to form the complete generative process.
At the token level, both the visual-conditional likelihood $P(y \mid v, x)$ and the language prior $P(y \mid x)$ admit autoregressive factorizations, meaning that the probability of the entire sequence is factorized into the product of the probabilities of each individual token conditioned on the previous ones:

\begin{equation}
\begin{aligned}
P(y \mid v, x) &= \prod_{t=1}^{T} P\!\big(y_t \mid y_{<t},\, v, x\big), \\
P(y \mid x)    &= \prod_{t=1}^{T} P\!\big(y_t \mid y_{<t},\, x\big),
\end{aligned}
\end{equation}

\noindent The first equation represents the likelihood of the output sequence given both the visual and textual context, while the second equation shows the likelihood given only the textual context. These factorizations treat each token in the sequence independently, conditioned on all previous tokens, allowing for sequential prediction.
To compare the visual-conditional likelihood and the language prior, we combine the joint factorizations with Bayes' rule. This results in a ratio that isolates the effect of the visual input relative to the language prior:

\begin{equation}
\frac{P(y \mid v, x)}{P(y \mid x)} = \frac{P(y, v \mid x)}{P(y \mid x)\,P(v \mid x)} = \frac{P(v \mid y, x)}{P(v \mid x)},
\end{equation}

\noindent This ratio quantifies how much the visual input $v$ contributes to the generation process beyond the language prior $P(y \mid x)$. The ratio serves as a \emph{PMI-style} measure of the influence of the visual context.
Taking the logarithm of this ratio turns it into a token-wise subtraction between the visual-conditional and the text-only next-token scores at each timestep $t$:

\begin{equation}
\begin{aligned}
\log \frac{P(y \mid v, x)}{P(y \mid x)} &= \sum_{t=1}^{T} \Big[\log P\!\big(y_t \mid y_{<t},\, v, x\big) \\
&\quad - \log P\!\big(y_t \mid y_{<t},\, x\big) \Big],
\end{aligned}
\end{equation}

\noindent This formulation allows us to evaluate how much the visual context adjusts the prediction for each token, removing the contribution of the language prior. Finally, the language prior itself can be expressed as a marginalization over the visual variable $v$, which accounts for all possible visual contexts. This marginalization is given by:

\begin{table}[t]
  \centering
  \setlength\tabcolsep{4pt}
  \fontsize{8}{9}\selectfont 
  
      \begin{tabular}{c|c|ccc}
            \toprule
            
            \multirow{2.5}{*}{Models} & \multirow{2.5}{*}{Historical Window $W$}
            & \multicolumn{2}{c}{POPE} & MME \\
            \cmidrule(lr){3-4} \cmidrule(lr){5-5}
            
            & & Accuracy $\uparrow$ & F1 $\uparrow$ & Accuracy $\uparrow$ \\
            \midrule
            \multirow{5}{*}{LLaVA-1.5}  
            & 2 & 87.05 & 86.59 & 1869.23 \\
            & 4 & 84.45 & 82.71 & 1762.76 \\
            & 8 & 87.23 & 86.93 & 1881.57 \\
            & 16 & 86.96 & 86.32 & 1882.41 \\
            & 32 & 87.35 & 87.03 & 1880.86 \\

            \midrule
            \multirow{5}{*}{Qwen2.5-VL}  
            & 2 & 84.33 & 85.21 & 2232.54 \\
            & 4 & 86.86 & 85.63 & 2326.93 \\
            & 8 & 90.16 & 89.56 & 2348.40 \\
            & 16 & 90.33 & 89.96 & 2350.52 \\
            & 32 & 89.98 & 89.73 & 2346.81 \\
            
            \bottomrule
        \end{tabular}
      \caption{\label{tab:window_w} 
      Performance of predefined historical window $W$ on LLaVA-1.5/Qwen2.5-VL for POPE and MME benchmarks.}
      \vspace{-0.2cm}
      
\end{table}

\begin{equation}
P(y \mid x) = \mathbb{E}_{v \sim P(v \mid x)}\!\big[P(y \mid v, x)\big],
\end{equation}
\noindent where the expectation is taken over all possible visual inputs $v$, conditioned on the textual context $x$. In practice, this expectation is approximated using Monte Carlo sampling, which gives:

\begin{equation}
P(y \mid x) \approx \frac{1}{n} \sum_{j=1}^{n} P\!\left(y \mid v_j, x\right),
\end{equation}
\noindent where $\{v_j\}_{j=1}^{n}$ are samples drawn from the image distribution conditioned on $x$. This Monte Carlo approximation effectively captures the language prior while accounting for the variability in the visual context.

\subsection{Theoretical Analysis}
\label{appendix:theoretical_analysis}


\paragraph{Definitions.}
Let $V$ be the visual variable, $H$ the history (image tokens + past text),
and $Y$ the next-token output.
Denote the baseline decoder by $p(y\mid v,h)$ and the residual-induced distribution by $r(y\mid v,h)$.
ResDec forms the $\alpha$-blended channel
\begin{equation*}
\begin{aligned}
p_\alpha(y\mid v,h)
&:= \frac{\,p(y\mid v,h)^{\,1-\alpha}\, r(y\mid v,h)^{\,\alpha}\,}
{\displaystyle Z_\alpha(v,h)} ,
\\[-2pt]
Z_\alpha(v,h)
&:= \sum_{y'\!}  p(y'\!\mid v,h)^{\,1-\alpha}  r(y'\!\mid v,h)^{\,\alpha}.
\end{aligned}
\end{equation*}
Write the $H$-marginal as
\begin{equation*}
\begin{aligned}
p_\alpha(y\mid h)
:= \sum_{v} p_\alpha(y\mid v,h)\,p(v\mid h).
\end{aligned}
\end{equation*}
The conditional mutual information is
\begin{equation*}
\begin{aligned}
I_\alpha(V;Y\mid H)
:= \mathbb{E}_{h}\!\left[
\mathrm{KL}\!\big(p_\alpha(\cdot\mid v,h) \,\|\, p_\alpha(\cdot\mid h)\big)
\right].
\end{aligned}
\end{equation*}
Define the \emph{residual advantage}
\begin{equation*}
\begin{aligned}
A(y,v,h)
:= \log r(y\mid v,h) - \log r(y\mid h).
\end{aligned}
\end{equation*}

\paragraph{Theorem 1.}
Assume regularity so that differentiation under the expectation is valid,
and suppose the residual is informative in the sense that for some $\varepsilon>0$,
\begin{equation*}
\begin{aligned}
\mathbb{E}_{h}\mathbb{E}_{v\mid h}\mathbb{E}_{y\sim p(\cdot\mid v,h)}
\!\left[A(y,v,h)\right] \ge \varepsilon .
\end{aligned}
\end{equation*}
Then the directional derivative of the conditional MI at $\alpha=0$ satisfies
{\small
\begin{equation*}
\begin{aligned}
\left.\frac{d}{d\alpha}\,I_\alpha(V;Y\mid H)\right|_{\alpha=0}
= \mathbb{E}_{h}\mathbb{E}_{v\mid h}\mathbb{E}_{y\sim p(\cdot\mid v,h)}
\!\left[A(y,v,h)\right] \ge \varepsilon .
\end{aligned}
\end{equation*}
}

\noindent Consequently, there exists $\alpha_0>0$ such that for all $\alpha\in(0,\alpha_0]$,
\begin{equation*}
\begin{aligned}
I_\alpha(V;Y\mid H) \ge  I_0(V;Y\mid H) + \tfrac{\varepsilon}{2}\,\alpha.
\end{aligned}
\end{equation*}

\paragraph{Proof.}
Rewrite $p_\alpha$ as an exponential tilt of $p$:
\begin{equation*}
\begin{aligned}
p_\alpha(y\mid v,h)
&= \frac{ p(y\mid v,h)\exp\!\big(\alpha\,U_v(y)\big) }{ Z_\alpha(v,h) }, \\
U_v(y)&:=\log r(y\mid v,h)-\log p(y\mid v,h).
\end{aligned}
\end{equation*}
Similarly,
\begin{equation*}
\begin{aligned}
p_\alpha(y\mid h)
&= \frac{ p(y\mid h)\exp\!\big(\alpha\,\bar U(y)\big) }{ \bar Z_\alpha(h) }, \\
\bar U(y)&:=\log r(y\mid h)-\log p(y\mid h).
\end{aligned}
\end{equation*}
For the log-densities, exponential-family calculus gives
\begin{equation*}
\begin{aligned}
\left.\frac{\partial}{\partial\alpha}\log p_\alpha(y\mid v,h)\right|_{0}
&= U_v(y)-\mathbb{E}_{y'\sim p(\cdot\mid v,h)}\!\left[U_v(y')\right],\\
\left.\frac{\partial}{\partial\alpha}\log p_\alpha(y\mid h)\right|_{0}
&= \bar U(y)-\mathbb{E}_{y'\sim p(\cdot\mid h)}\!\left[\bar U(y')\right].
\end{aligned}
\end{equation*}
\noindent Differentiating the KL form of $I_\alpha$ at $\alpha=0$ and using
$\frac{d}{d\alpha}\mathrm{KL}(P_\alpha\|Q_\alpha)
= \mathbb{E}_{P_\alpha}\!\big[\partial_\alpha\log P_\alpha
-\partial_\alpha\log Q_\alpha\big]$ at $\alpha=0$ yields
\begin{equation*}
\begin{aligned}
\left.\frac{d}{d\alpha}\,I_\alpha\right|_{0}
= \mathbb{E}_{h}\mathbb{E}_{v\mid h}\mathbb{E}_{y\sim p(\cdot\mid v,h)}
\!\Big[
\big(U_v(y)-\bar U(y)\big)
\Big].
\end{aligned}
\end{equation*}
With the definitions of $U_v$ and $\bar U$, the bracket equals
\begin{equation*}
\begin{aligned}
&\big(\log r(y\mid v,h) -\log p(y\mid v,h)\big) \\
&-\big(\log r(y\mid h)-\log p(y\mid h)\big) \\
&= A(y,v,h) - \mathrm{PMI}_p(y;v\mid h).
\end{aligned}
\end{equation*}
Taking expectation over $y\sim p(\cdot\mid v,h)$ cancels the $\mathrm{PMI}_p$
term, giving the stated derivative. The claimed local increase follows by continuity.


\paragraph{Theorem 2.}
Let $v$ be the visual input, $x$ the text query, and $h$ the decoding history.
Denote the base logits by $\ell(y\,|\,v,x,h)$ and the residual offsets by $r_h(y)$.
Define the base and residual-augmented conditional distributions
\begin{equation*}
p_0(y\,|\,v,x,h)  =  \mathrm{softmax}\!\big(\ell(y\,|\,v,x,h)\big),
\end{equation*}
\begin{equation*}
p_\alpha(y\,|\,v,x,h)  =  \mathrm{softmax}\!\big(\ell(y\,|\,v,x,h)+\alpha\,r_h(y)\big),
\end{equation*}
with $\alpha\in(0,1]$. Equivalently,
\begin{equation*}
\begin{aligned}
p_\alpha(y\,|\,v,x,h)
= \frac{p_0(y\,|\,v,x,h) \exp\!\big(\alpha\,r_h(y)\big)}
{\displaystyle Z(\alpha)}, \\
Z(\alpha)=\sum_{y'} p_0(y'\,|\,v,x,h)\,\exp\!\big(\alpha\,r_h(y')\big).
\end{aligned}
\end{equation*}
Assume the residual aligns with the desirable visual grounding in the sense that
$\operatorname{Cov}_{p_\alpha}\!\big(r_h(Y),\,\log p_0(Y\,|\,v,x,h)\big)\ge 0$
for $\alpha\in[0,1]$, and $\operatorname{Var}_{p_\alpha}\!\big(r_h(Y)\big)>0$.
Then, for any $\alpha\in(0,1]$, the conditional Shannon entropy strictly decreases:
\begin{equation*}
H\!\big(Y\,\big|\,v,x,h\big)_{p_\alpha}
< H\!\big(Y\,\big|\,v,x,h\big)_{p_0}.
\end{equation*}

\paragraph{Proof.}
Use the exponential-tilted form of $p_\alpha$ displayed above and write the conditional
entropy under $p_\alpha$ as
\begin{equation*}
\begin{aligned}
H_\alpha &= -\sum_{y} p_\alpha(y)\,\log p_\alpha(y) \\
&= -\mathbb{E}_{p_\alpha}\!\big[\log p_0(Y)\big]
- \alpha\,\mathbb{E}_{p_\alpha}\!\big[r_h(Y)\big]
+\log Z(\alpha).
\end{aligned}
\end{equation*}

\noindent Differentiate with respect to $\alpha$.
Invoking the standard score function identity for the $\alpha$-indexed family $\{p_\alpha\}$,
namely
$\frac{\mathrm{d}}{\mathrm{d}\alpha}\,\mathbb{E}_{Y\sim p_\alpha}\!\big[f(Y)\big]
=\operatorname{Cov}_{Y\sim p_\alpha}\!\big(f(Y),\,r_h(Y)\big)$
for any integrable test function $f$, and using
$\frac{\mathrm{d}}{\mathrm{d}\alpha}\,\log Z(\alpha)
=\mathbb{E}_{Y\sim p_\alpha}\!\big[r_h(Y)\big]$, we obtain
\begin{equation*}
\begin{aligned}
\frac{\mathrm{d}}{\mathrm{d}\alpha}H_\alpha
&= -\,\operatorname{Cov}_{p_\alpha}\!\big(\log p_0(Y),\,r_h(Y)\big)
 - \mathbb{E}_{p_\alpha}\!\big[r_h(Y)\big] \\
 &\phantom{=} - \alpha\,\operatorname{Cov}_{p_\alpha}\!\big(r_h(Y),\,r_h(Y)\big)
 + \mathbb{E}_{p_\alpha}\!\big[r_h(Y)\big].
\end{aligned}
\end{equation*}
The middle expectations cancel, yielding
{\small
\begin{equation*}
\frac{\mathrm{d}}{\mathrm{d}\alpha}H_\alpha
 =  -\,\operatorname{Cov}_{p_\alpha}\!\big(\log p_0(Y),\,r_h(Y)\big)
 - \alpha\,\operatorname{Var}_{p_\alpha}\!\big(r_h(Y)\big).
\end{equation*}
}

\noindent Under the alignment assumption, $r_h(Y)$ is positively associated with the base
log-likelihood $\log p_0(Y)$, hence
$\operatorname{Cov}_{p_\alpha}\!\big(\log p_0(Y),\,r_h(Y)\big)\ge 0$.
Moreover, the residual signal is non-degenerate under $p_\alpha$, so
$\operatorname{Var}_{p_\alpha}\!\big(r_h(Y)\big)>0$.
These conditions hold for all $\alpha$ and together capture positive association with
nontrivial variability; therefore, for any $\alpha>0$,
\begin{equation*}
\frac{\mathrm{d}}{\mathrm{d}\alpha}H_\alpha  <  0.
\end{equation*}
Integrating from $0$ to any $\alpha\in(0,1]$ gives
\begin{equation*}
H_\alpha - H_0  =  \int_{0}^{\alpha} \frac{\mathrm{d}}{\mathrm{d}s}H_s\,\mathrm{d}s  <  0,
\end{equation*}
which is precisely the claimed entropy decrease.


\section{Additional Ablation Studies}
\subsection{Predefined Historical Window}
\label{appendix:window}
In this section, we conduct experiments on the Predefined Historical Window $W$. Since $W$ merely serves as a parameter for identifying the U-shape, we set $W$ to 2, 4, 8, 16, and 32. The corresponding experimental results are presented in Table~\ref{tab:window_w}. We find that the U-shape is typically achievable when $W=8$; thus, we usually set $W=8$ in our experiments. 

\subsection{Historical Window Selection Strategies}
As discussed in Sec.~3.2, the temporal evolution of candidate tokens can be decomposed into three phases: the Pre-Semantic Clarity Phase (PSAP), the Semantic Anchoring Phase (SAP), and the Expressive Divergence Phase (EDP). To validate the necessity of leveraging both SAP and EDP for historical information aggregation, we ablate the window selection on Qwen2.5-VL and LLaVA-1.5 using the POPE and MME benchmarks. 

We compare our \textit{ResDec} strategy against using the Full Window (PSAP+SAP+EDP), individual phases (PSAP only, SAP only, EDP only), and a heuristic Top-4 Confident Steps selection. As shown in Table~\ref{tab:ablation_window_selection}, incorporating the PSAP degrades performance. This degradation is likely due to the candidate token distribution transitioning from an initial state of disorder, which introduces early noise before settling on core semantics. While relying solely on the SAP improves accuracy, it slightly trails the full \textit{ResDec} strategy. This confirms our observation that incorporating the EDP---which contains effective decoding guidance due to its diverse expressions---provides vital contextual information for maintaining coherent generation.

\begin{table}[th]
  \centering
  \setlength\tabcolsep{4pt}
  \fontsize{7}{6}\selectfont 
  \resizebox{\linewidth}{!}{
      \begin{tabular}{c|c|ccccc}
            \toprule
            \toprule
            \multirow{2.6}{*}{Models} & \multirow{2.6}{*}{Strategy} & 
            \multicolumn{2}{c}{POPE} & \multicolumn{1}{c}{MME} 
            \\
            \cmidrule(lr){3-4} \cmidrule(lr){5-5} 
            & & ACC $\uparrow$ & F1 $\uparrow$ & ACC $\uparrow$ \\
            \midrule
            \multirow{6}{*}{$\text{Qwen2.5-VL }$} 
            & Full Window (PSAP+SAP+EDP) & 85.83  & 84.72 & 2275.68 \\
            & PSAP Only & 85.17  & 84.24 & 2151.31 \\
            & SAP Only  & 88.82  & 87.96   & 2337.68 \\
            & EDP Only  & 86.14   & 84.27 & 2330.47 \\
            & Top-4 Confident Steps  & 82.91  & 79.63 & 2142.87 \\
            & \textit{ResDec} (ours) & \textbf{90.16} & \textbf{89.56} & \textbf{2348.40} \\

            \midrule 
            \multirow{6}{*}{$\text{LLaVA-1.5}$}
            & Full Window (PSAP+SAP+EDP) & 84.87   & 84.03 & 1813.62   \\
            & PSAP Only & 84.13   & 83.32 & 1797.94 \\
            & SAP Only  & 86.21   & 85.87 & 1846.45 \\
            & EDP Only  & 85.56  & 84.74  & 1843.32 \\
            & Top-4 Confident Steps  & 81.21   & 79.46  & 1743.78  \\
            & \textit{ResDec} (ours) & \textbf{87.23} & \textbf{86.93} & \textbf{1881.57} \\

            \bottomrule
            \bottomrule
        \end{tabular}
        }
      \caption{\label{tab:ablation_window_selection}
      Ablation study on historical window selection strategies.
      }
\end{table}

\subsection{Aggregation Strategy}

\textit{ResDec} employs a confidence-weighted aggregation method based on a local confidence metric derived from the model's internal token distribution. To justify this proposed pooling strategy, we compare it against two alternative heuristic aggregation methods: Uniform (Mean) Pooling and Distance Decay (inverse temporal-distance weighting).

The results presented in Table~\ref{tab:aggregation_strategy} demonstrate that \textit{ResDec}'s confidence-weighted pooling consistently outperforms both uniform and distance-based decay strategies. By dynamically weighting historical time steps based on their degree of certainty, \textit{ResDec} more effectively isolates stable semantics and mitigates language-prior hallucinations compared to fixed or strictly distance-based heuristic weightings.

\begin{table}[th]
  \centering
  \setlength\tabcolsep{4pt}
  \fontsize{7}{6}\selectfont 
  \resizebox{\linewidth}{!}{
      \begin{tabular}{c|c|ccccc}
            \toprule
            \toprule
            \multirow{2.6}{*}{Models} & \multirow{2.6}{*}{Aggregation Strategy} & 
            \multicolumn{2}{c}{POPE} & \multicolumn{1}{c}{MME} 
            \\
            \cmidrule(lr){3-4} \cmidrule(lr){5-5} 
            & & ACC $\uparrow$ & F1 $\uparrow$ & ACC $\uparrow$ \\
            \midrule
            \multirow{3}{*}{$\text{Qwen2.5-VL }$} 
            & Uniform (Mean) Pooling & 88.31  & 87.21  & 2342.76  \\
            & Distance Decay  & 86.93  & 85.45 & 1954.79  \\
            & \textit{ResDec} (ours) & \textbf{90.16} & \textbf{89.56} & \textbf{2348.40} \\

            \midrule 
            \multirow{3}{*}{$\text{LLaVA-1.5}$}
            & Uniform (Mean) Pooling & 85.26  & 84.74   & 1785.41  \\
            & Distance Decay  & 83.21 & 82.55  & 1731.62  \\
            & \textit{ResDec} (ours) & \textbf{87.23} & \textbf{86.93} & \textbf{1881.57} \\

            \bottomrule
            \bottomrule
        \end{tabular}
        }
      \caption{\label{tab:aggregation_strategy}
        Ablation study on the aggregation strategy.
      }
\end{table}

\section{More Case Studies}
\label{appendix:case_study}
Additional case studies are presented
to illustrate the effectiveness of our \textit{ResDec}.

\begin{figure*}[th]
  \centering
  \includegraphics[width=1.0\linewidth, page=1]{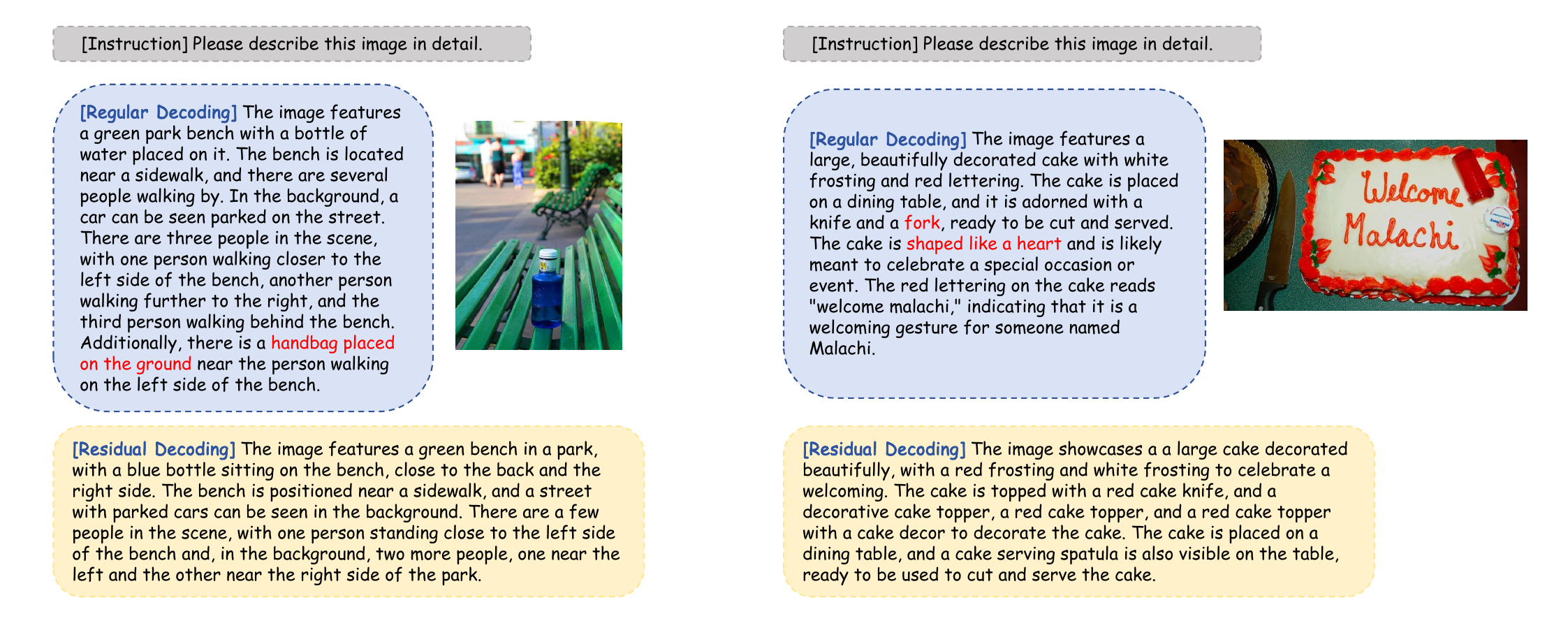}
  \vspace{0.3cm}  
  \includegraphics[width=1.0\linewidth, page=2]{imgs/case_study234.pdf}
  \vspace{0.3cm}  
  \includegraphics[width=1.0\linewidth, page=3]{imgs/case_study234.pdf}
  
  \caption{More examples comparing regular decoding and residual decoding in image captioning. Hallucinated details in ``regular'' decoding outputs are highlighted in \textcolor{red}{red}.}
  \label{fig:case234}
  \vspace{-0.2cm}
\end{figure*}

\begin{figure*}[th]
  \centering
  \includegraphics[width=1.0\linewidth, page=1]{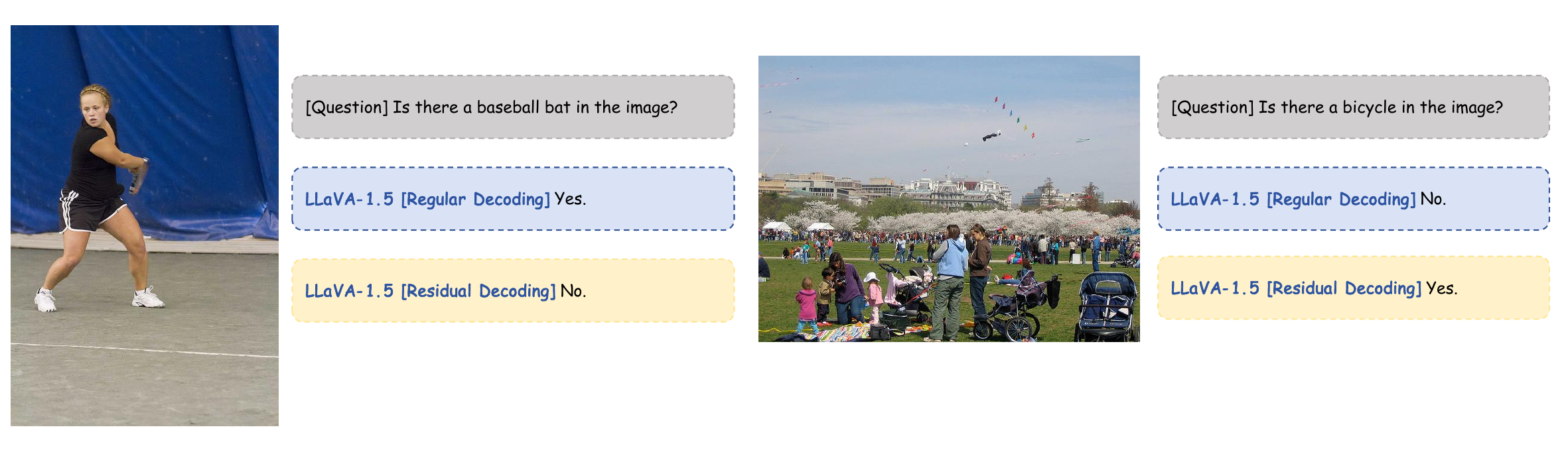}
  \vspace{0.3cm}  
  \includegraphics[width=1.0\linewidth, page=2]{imgs/case_study_2.pdf}
  \vspace{0.3cm}  
  \includegraphics[width=1.0\linewidth, page=3]{imgs/case_study_2.pdf}
  
  \caption{More examples comparing regular decoding and residual decoding in image captioning. Hallucinated details in ``regular'' decoding outputs are highlighted in \textcolor{red}{red}.}
  \label{fig:case_2}
  \vspace{-0.2cm}
\end{figure*}

\begin{figure*}[th]
  \centering
  \includegraphics[width=1.0\linewidth, page=1]{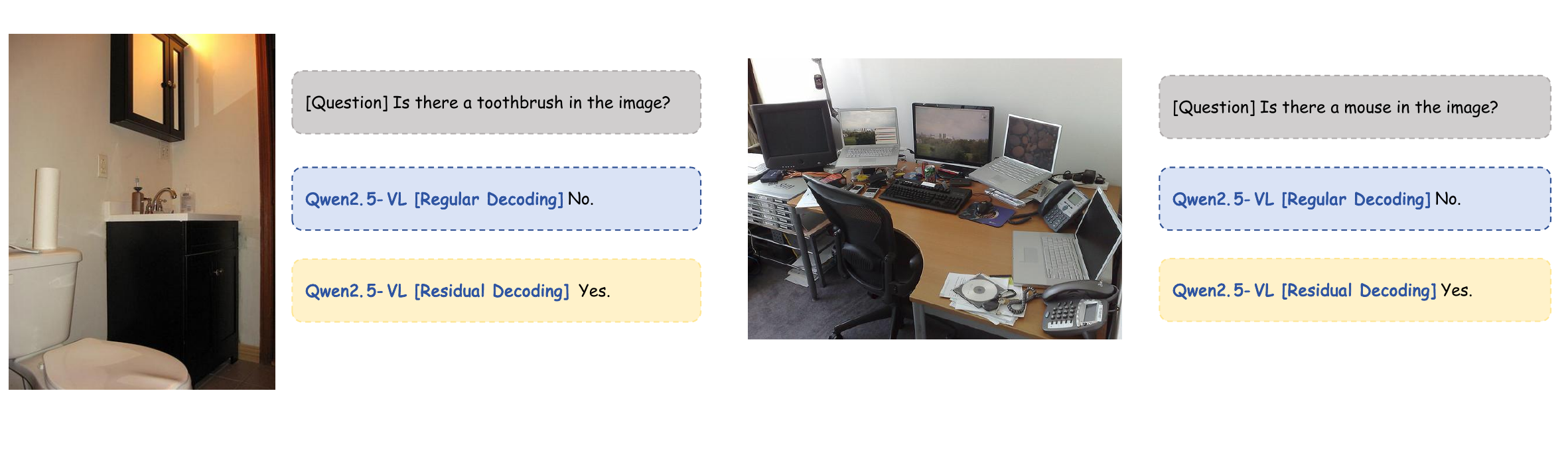}
  \vspace{0.3cm}  
  \includegraphics[width=1.0\linewidth, page=2]{imgs/case_study_3.pdf}
  \vspace{0.3cm}  
  \includegraphics[width=1.0\linewidth, page=3]{imgs/case_study_3.pdf}
  
  \caption{More examples comparing regular decoding and residual decoding in image captioning. Hallucinated details in ``regular'' decoding outputs are highlighted in \textcolor{red}{red}.}
  \label{fig:case_3}
  \vspace{-0.2cm}
\end{figure*}


\end{document}